\documentclass{article}
\usepackage{tikz}
\usetikzlibrary{positioning}
\usepackage{amsmath}
\usepackage[final]{corl_2026} 
\usepackage{amsfonts}
\usepackage{multirow}
\usepackage{array}

\title{-}

\usepackage{wrapfig}

\title{Mind the Phase: Effective Rank and Representation Health in Legged Locomotion}

\newcommand{\affilmark}[1]{\textsuperscript{\normalfont #1}}

\author{
    Felipe Tommaselli\affilmark{*}
    \And
    Ricardo V. Godoy
    \And
    Thiago H. Segreto
    \AND
    Juliano D. Negri
    \And
    Marcelo Becker
    \AND
    \normalfont
    \textbf{Mobile Robotics Group, University of S\~ao Paulo}\\
    \affilmark{*}Corresponding author. Email: f.tommaselli@usp.br 
} 

\begin{document}
\maketitle
\vspace{-0.18in}

\begin{figure}[h]
    \centering
    \includegraphics[width=0.9\linewidth]{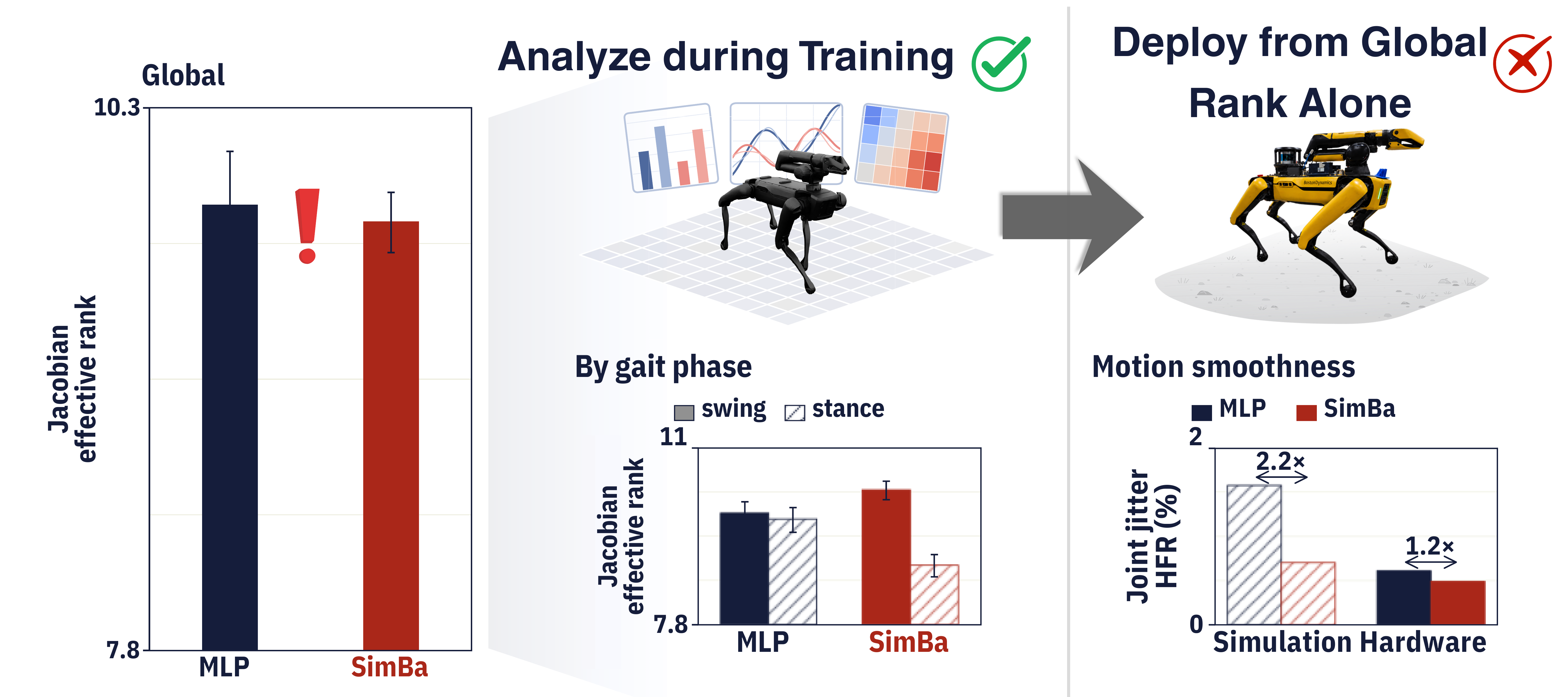}
    \caption{\textbf{Global rank leaves gait-phase structure invisible.} How a policy distributes its capacity across the gait cycle is associated with its health and motion smoothness, both in simulation and on hardware. Website: \href{https://felipe-tommaselli.github.io/mindthephase/}{\nolinkurl{felipe-tommaselli.github.io/mindthephase}}}
    \label{fig:spot}
\end{figure}


\begin{abstract} Reinforcement learning has become the leading paradigm in legged locomotion, enabling complex behaviors from backflips to parkour via massively parallel simulation. Under PPO's non-stationarity, shallow networks remain the de facto architecture, supported by carefully staged environments; nevertheless, the representations learned by these policies remain poorly understood, specifically the relation to motion produced on hardware. In this work, we empirically study locomotion policies through the effective rank of the network Jacobian and show that conditioning rank on gait phase exposes representation structures invisible from a global rank. In particular, we find that architectural choices, namely observation and layer normalization with residual connections, allocate more than one additional dimension of effective rank to swing-dominant states than to stance-dominant states, which is mostly absent in vanilla MLPs. Building on this, we propose practical recommendations to convert these into diagnostics for healthier and smoother policies. In practice, this results in lower joint jitter, up to $2$x, not only in simulation, but also on a physical Spot. More broadly, our findings show that how a policy distributes its capacity across the gait cycle is closely tied to its health and sim-to-real transfer.

\end{abstract}


\section{Introduction}
\label{sec:intro}

Deep Reinforcement Learning (RL) has been the backbone for recent progress in robotics agility, unlocking previously unreachable acrobatic and athletic behaviors \citep{doi:10.1126/scirobotics.aau5872, zhuang2023robotparkourlearning, he2025attentionbasedmapencodinglearning}. The combination of scalable on-policy RL with a fast parallelizable simulator has enabled zero-shot sim-to-real policies with highly dynamic capabilities \citep{rudin2022learning}. Much of this progress has been supported by advances in domain randomization and system identification~\citep{tan2018sim}, curriculum learning~\citep{rudin2022learning}, policy distillation~\citep{doi:10.1126/scirobotics.abk2822}, mixture-of-experts policies~\citep{affonso2026ctsmoe} and more. In such cases, continual learning in increasingly difficult conditions places sustained demands on the network's ability to adapt throughout training.

Under this regime, preserving representational health is a persistent challenge. Proximal Policy Optimization (PPO)~\cite{schulmanProximalPolicyOptimization2017} remains the most widely used algorithm for on-policy RL, despite its well-known vulnerability to non-stationarity and trust-region collapses. In the non-stationary objective of PPO, the network can lose its capacity to fit new targets~\citep{pmlr-v232-abbas23a, moalla2024no}, making this continued adaptation a concern as training progresses.

Layer normalization and residual connections have supported more stable learning and allowed larger networks to remain effective in deep RL~\citep{lee2025simba}. For legged locomotion, however, preserving representational capacity leaves a further question: how do the representations learned during training relate to the behavior that the policy produces? Policies that achieve comparable returns may still differ in how they respond to observations and in the motion they produce, with practical consequences for deployment on hardware. Understanding these differences would help connect architectural choices and training dynamics to aspects of control that are not fully described by aggregate task performance.

Locomotion provides a natural setting in which to examine this question because the demands on the controller change throughout the gait cycle. Supporting the body during stance and moving a foot toward its next contact during swing are parts of the same behavior~\citep{raibert1986legged}, but do not necessarily elicit the same response from the policy. In legged locomotion, the field has largely converged on shallow MLPs supervised almost entirely by meticulously tuned reward signals, leaving minimal traces on the induced representations of these networks. 

In this work, we take a mechanistic view of these questions and study locomotion policies by means of the effective rank of the policy Jacobian, using rank as a diagnostic of training health. Because the Jacobian describes how small changes in observations affect actions, it provides a way to examine the policy's sensitivity and investigate its relationship with motion smoothness. We examine how its effective rank varies with gait phase, architecture, and training regime, while separately measuring the smoothness of the resulting motion.

Interestingly, we found that due to the nature of legged gait cycles, a single scalar for global rank is misleading. Healthier policies consistently across multiple embodiments show a delta between the representation allocated for the swing and stance phases. We further examine how this correlates with motion quality in simulation and on Boston Dynamics Spot. 

Concretely, we summarize the following contributions:
\begin{itemize}
    \item \textbf{A phase-conditioned rank diagnostic.} We show that conditioning effective Jacobian rank on swing-dominant and stance-dominant states reveals differences between Spot locomotion policies that are invisible for a global rank.
    \item \textbf{An empirical study of architecture and training effects.} We characterize how rank structure differs between MLP and SimBa policies and changes under increased PPO updates, showing why rank magnitude alone is insufficient to assess representation health.
    \item \textbf{A connection to deployment quality.} We examine the association between rank and joint-motion smoothness, with comparisons across three simulated embodiments and deployment on a physical Spot, motivating their joint use in policy evaluation.
\end{itemize}


\section{Related Works}
\label{sec:rw}

\textbf{Deep RL for legged locomotion.}
A recurring characteristic of modern legged locomotion is the use of relatively simple policy architectures within increasingly sophisticated learning pipelines. Early sim-to-real work demonstrated neural locomotion policies on ANYmal~\citep{doi:10.1126/scirobotics.aau5872}, while \citet{doi:10.1126/scirobotics.abc5986} extended this paradigm to blind locomotion over challenging natural terrain through a teacher-student scheme. Combined with massively parallel simulation~\citep{rudin2022learning, mittal2023orbit}, similar pipelines now produce highly athletic behaviors, including the diverse parkour skills of \citet{zhuang2023robotparkourlearning} and the extreme terrain traversal of \citet{cheng2023extremeparkourleggedrobots}. Much of the structure required for learning is introduced through rewards, curricula, observations, reference motions, or privileged supervision, while the low-level policy itself is often kept compact. The gait structure is a particularly important example: classical controllers explicitly organize locomotion around foot placement and alternating swing and stance phases~\citep{raibert1986legged}, while learned controllers may instead recover this periodic structure from interaction or from phase-dependent objectives~\citep{siekmann2021sim}. At deployment, a related concern is that reward-optimized policies can exploit high-frequency control signals that are undesirable or unrealizable on hardware. A common approach is therefore to penalize action rates through reward shaping~\citep{mysore2021regularizingactionpoliciessmooth}. More recent work regularizes the policy input-output map directly: \citet{chen2025learning} constrains policy sensitivity through a gradient-based Lipschitz penalty, while \citet{xie2026learning} penalizes the full action Jacobian and relates the two formulations. These works use policy derivatives as regularization objectives; here, we instead study the Jacobian as a representation-level diagnostic and examine how its structure varies across gait phases.

\textbf{Deep networks under PPO's non-stationarity.} PPO's clipped ratio~\citep{schulmanProximalPolicyOptimization2017} limits how far each update can move the action distribution, but not the underlying representations. In practice, it's known that this ratio can escape the clip interval ~\citep{engstrom2019implementation}, allowing the underlying representations to drift under PPO's non-stacionarity ~\citep{igl2020transient}. This cathastrophical behavior is often related to the degradation of the feature space~\citep{kumar2020implicit}, which becomes rank-deficient that leads to plasticity loss and the network inability to fit new targets ~\citep{dohare2021continual, pmlr-v202-lyle23b}. \citet{moalla2024no} extends this from value-based literature to PPO, showing that feature rank collapse and capacity loss accompany a degradation of the trust region. The right architectural priors keep high-capacity networks trainable, through residual blocks and normalization in SimBa~\citep{lee2025simba}, hyperspherical normalization in SimBaV2~\citep{lee2025hyperspherical}, and bounded norms in FlashSAC~\citep{kim2026flashsac}. 


\section{Background}
\label{sec:background}

\subsection{Reinforcement Learning}

We formalize our RL setting as a discounted Markov decision process with states $\mathcal{S}$, actions $\mathcal{A}$, reward $r(s,a,s')$~\citep{ReinforcementLearningIntroductionsutton}; the policy acts on an observation $o_t$ of $s_t$, which we write as $s_t$ when no confusion arises. At each step $t$ the agent observes $s_t$, samples $a_t \sim \pi_\theta(\cdot \mid s_t)$ from a Gaussian policy, transitions to $s_{t+1}$, and receives $r_{t+1}$, seeking to maximize the expected discounted return $J(\pi_\theta) = \mathbb{E}_{\pi_\theta}\!\left[\sum_{t=0}^{H-1} \gamma^t r_{t+1}\right]$ over episodes of horizon $H$. Training data are collected in shorter rollouts of $T$ steps per environment.

In the on-policy actor-critic pipeline of \citet{rudin2022learning}, a policy network $\pi_\theta$ and a value network $\hat{v}_w$ are optimized jointly. Each iteration collects a rollout with the current $\pi_{\mathrm{old}}$, computes advantages $\hat{A}_t$ with GAE \citep{schulman2015high}, and forms return targets $\hat{R}_t = \hat{A}_t + \hat{v}_{w_{\mathrm{old}}}(s_t)$. PPO-Clip \citep{schulmanProximalPolicyOptimization2017} then maximizes:
\begin{equation}
    L^{\mathrm{CLIP}}(\theta) =
    \hat{\mathbb{E}}_t\!\left[
        \min\!\left(
            \rho_t \hat{A}_t,\;
            \mathrm{clip}(\rho_t,\, 1-\varepsilon,\, 1+\varepsilon)\,
            \hat{A}_t
        \right)
    \right],
    \qquad
    \rho_t = \frac{\pi_\theta(a_t \mid s_t)}{\pi_{\mathrm{old}}(a_t \mid s_t)},
\end{equation}
where $\hat{\mathbb{E}}_t$ denotes the empirical mean over the transitions of a minibatch. The critic minimizes the clipped value loss
$\mathcal{L}^V(w) = \hat{\mathbb{E}}_t\big[\max\big((\hat{v}_w(s_t) - \hat{R}_t)^2,\, (\hat{v}^{\mathrm{clip}}_w(s_t) -
\hat{R}_t)^2\big)\big]$, where $\hat{v}^{\mathrm{clip}}_w(s_t)$ is $\hat{v}_w(s_t)$ clipped to within $\varepsilon$ of
$\hat{v}_{w_{\mathrm{old}}}(s_t)$. The agent optimizes the joint loss over $K$ minibatch epochs:
\begin{equation}
    \mathcal{L}(\theta, w) =
    -L^{\mathrm{CLIP}}(\theta)
    + \underbrace{c_v\,\mathcal{L}^V(w)}_{\text{weighted value loss}}
    - \underbrace{c_H\,\hat{\mathbb{E}}_t\!\left[
        \mathcal{H}\!\left[\pi_\theta(\cdot \mid s_t)\right]
    \right]}_{\text{weighted entropy bonus}}.
\end{equation}

\subsection{Effective Rank}

\begin{wrapfigure}{r}{0.45\columnwidth}
  \centering
  \vspace{-1em}
  \includegraphics[width=0.38\columnwidth]{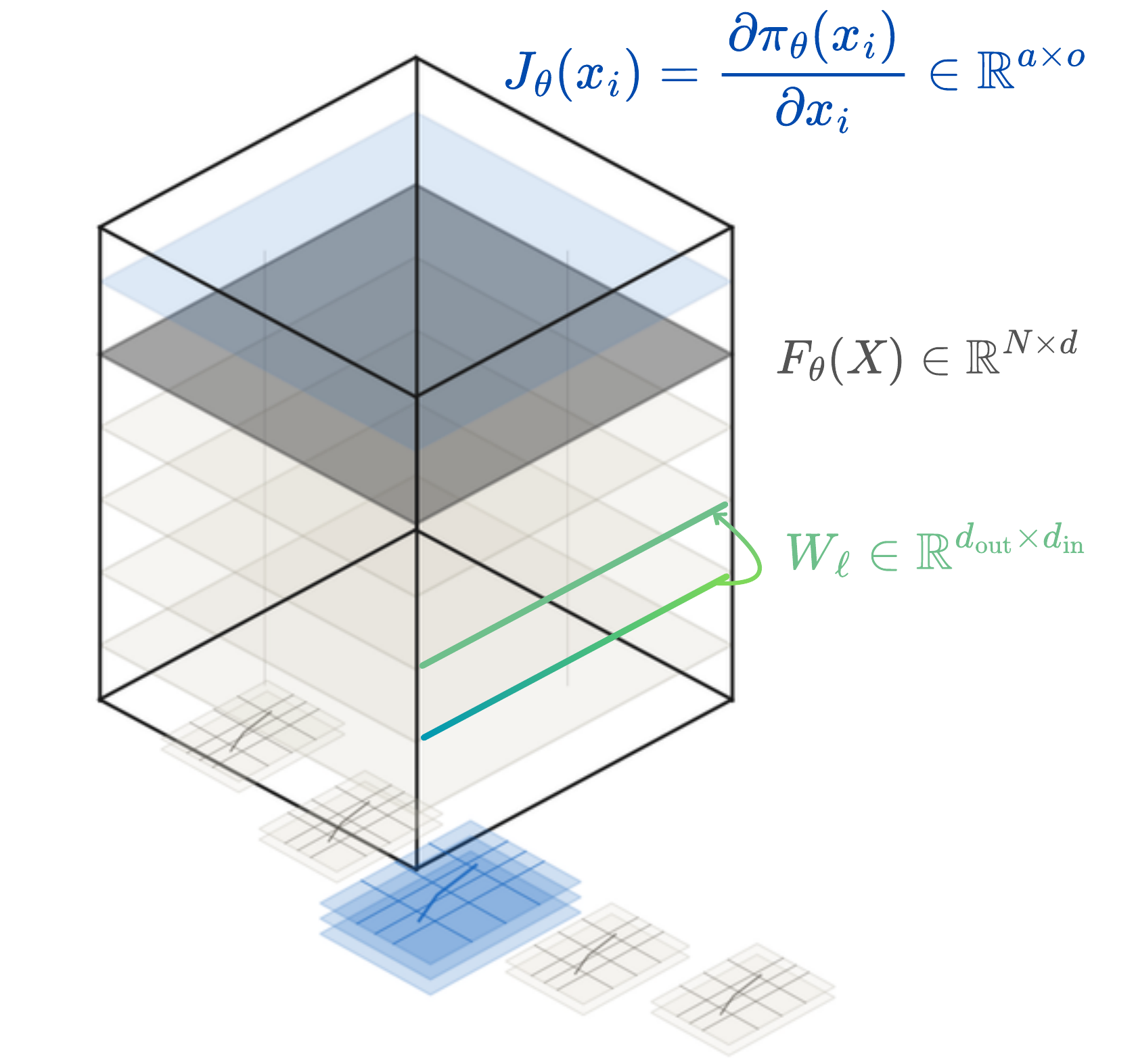}
\caption{\textbf{Overview of the rank objects} studied here, namely the actor weight matrices $W_\ell$, the penultimate feature representation $F_\theta(X)$, and the policy Jacobian $J_\theta(x_i)$ measuring local input-output sensitivity.}
  \label{fig:rank_diagram}
  \vspace{-1em}
\end{wrapfigure}

In this work, we adopt several rank formulations used in the broader deep learning literature. We characterize the policy through the entropy-based effective rank \citep{roy2007effective} of the analyzed activation matrix (Fig.~\ref{fig:rank_diagram}). For a matrix $A$ with singular values $\{\sigma_i\}$, we discard numerically negligible modes $\sigma_i \leq \xi$, normalize the remaining spectrum as $p_i = \sigma_i / \sum_j \sigma_j$, and define
\begin{equation}
    \operatorname{erank}(A)
    =
    \exp\left(
        - \sum_i p_i \log(p_i + \xi)
    \right),
    \label{eq:effective_rank}
\end{equation}
where the small constant $\xi$ also stabilizes the logarithm. This quantity can be interpreted as the entropy-equivalent number of active spectral modes \citep{roy2007effective, huh2021low}. A flat spectrum over $k$ modes gives $\operatorname{erank}(A) \approx k$, whereas a spectrum dominated by a single singular value gives $\operatorname{erank}(A) \approx 1$.

\paragraph{Feature rank.}

Feature rank is the most common rank in deep RL \citep{kumar2020implicit, moalla2024no}. We evaluate the actor's features, the output of its second-to-last linear layer before the activation (Appendix~\ref{app:rank}), on a batch of $N$ observations from the parallel environments, stack them into $F_\theta(X) \in \mathbb{R}^{N \times d}$, and report $\operatorname{erank}(F_\theta)$. For qualitative inspection, we also form the cosine-similarity Gram matrix $G = \hat{F}\hat{F}^{\top}$ \citep{huh2021low} of the L2-normalized feature rows; its singular values are the squares of those of $\hat{F}$. Finally, we also compute effective ranks of the actor's linear weight matrices $W_\ell \in \mathbb{R}^{d_{\mathrm{out}} \times d_{\mathrm{in}}}$, termed Weight Rank. This can be interpreted as a non-representational parametric rank, which is useful for checking available capacity for downstream representations. For instance, a high representation rank ought to be uninformative if the weight rank has collapsed. Conversely, feature rank can collapse or not, even with a high weight rank.   

\paragraph{Policy Jacobian rank.}
We additionally consider the policy Jacobian effective rank, which measures the dimensionality of the policy's input-output sensitivity. For the deterministic action mean $\mu_\theta(x) \in \mathbb{R}^{a}$ and an observation $x_i \in \mathbb{R}^{o}$, the per-state Jacobian is $J_\theta(x_i) = \partial \mu_\theta(x_i) / \partial x_i \in \mathbb{R}^{a \times o}$, taken with respect to the network input. Averaging over a batch of $B$ states drawn from a fixed pool sampled at the start of training gives $\bar{J}_\theta = \frac{1}{B}\sum_{i=1}^{B} J_\theta(x_i)$, and we define $r_{\mathrm{Jac}} = \operatorname{erank}(\bar{J}_\theta)$, which is bounded by the action dimension $a$. A low value indicates that the policy actions depend on only a few dominant observation directions, while a higher value indicates broader sensitivity across observation dimensions and action coordinates.

To ask how this response changes with the robot's support condition, we apply the same construction separately to groups of states. For a group $p$ (Section~\ref{sec:gait_phase}), we draw $n$ states $\mathcal{I}_p$ from rollouts of a saved checkpoint and compute $\bar{J}_{\theta,p} = n^{-1}\sum_{i\in\mathcal{I}_p} J_\theta(x_i)$. The phase difference is $\Delta_\phi = \operatorname{erank}(\bar{J}_{\theta,\mathrm{swing}}) - \operatorname{erank}(\bar{J}_{\theta,\mathrm{stance}})$, and the global rank applies the same estimator to all states. Every group uses the same number of states, the same policy, and the same observation coordinates, so their ranks are directly comparable (Appendix~\ref{append:contact_conditioning}).


\subsection{Gait Phase}
\label{sec:gait_phase}

\begin{wrapfigure}{r}{0.45\columnwidth}
  \centering
  \vspace{-1em}
  \includegraphics[width=0.33\columnwidth]{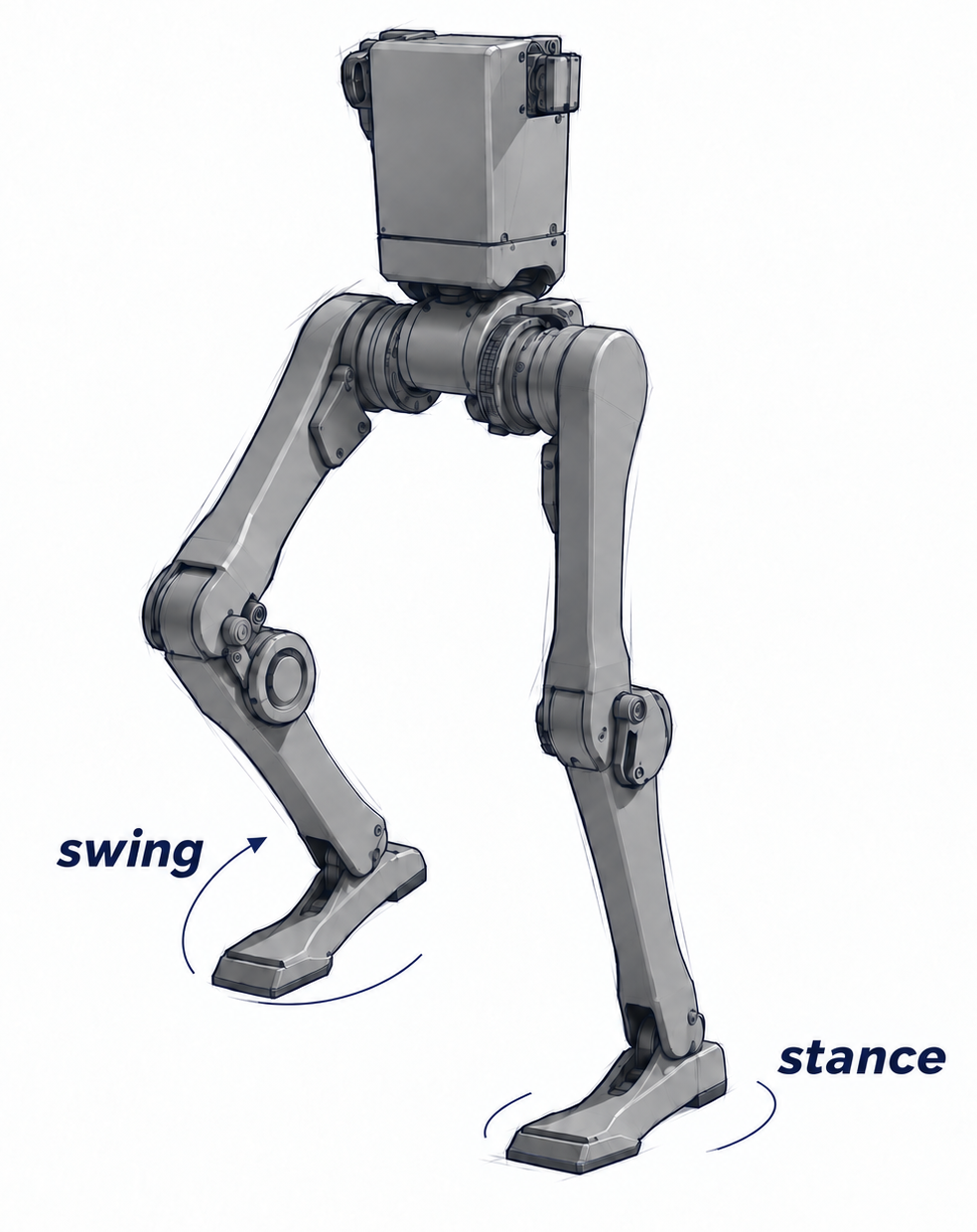}
  \caption{\textbf{Swing and stance phases during legged locomotion}. In stance, the foot remains in contact with the ground and supports the robot, while in swing it moves toward the next contact.}
  \label{fig:gait_phase}
  \vspace{-1em}
\end{wrapfigure}

Legged locomotion is naturally organized around alternating periods of \textit{stance}, when a foot is in contact with the ground, and \textit{swing}, when it moves toward the next contact~\citep{raibert1986legged}. These phases place different demands on the controller; note that stance is associated with supporting and stabilizing the body, while swing is primarily concerned with repositioning the leg for the following step. Figure~\ref{fig:gait_phase} illustrates this distinction for the Unitree H1.

Since a single policy controls the whole body (all legs) at once, we aggregate this information into groups. For a quadruped, states with zero or one foot down are \textit{swing-dominant}, and states with three or four feet down are \textit{stance-dominant}; double support is neutral and excluded from the aggregation to avoid inflating the count. These groups cover the full range of behavior the policy encounters in training, a mixture of velocity commands, including standing still (full stance). Contact criteria and sampling details are given in Appendix~\ref{append:contact_conditioning}.

A policy that responds very differently, or along many more Jacobian directions, in one support condition than in another may therefore also move its joints differently depending on how the body is supported. One consequence of this response that we can measure directly is jitter, by which we mean small and rapid oscillations of the joints that do not contribute to the commanded motion, and which on hardware can cause vibration on the actuators. We quantify jitter as the fraction of the joint-position signal energy above $10$\,Hz, adapted from~\citep{xie2026learning}. Full formulation is available in Appendix~\ref{app:simeval}.

\section{Experiments}
\label{sec:experiments}

We begin our experiments by evaluating PPO configurations using RSL\_RL \cite{rudin2022learning} within IsaacLab \cite{mittal2025isaaclab}, a standard platform for legged locomotion learning (Fig.~\ref{fig:spot}). We follow an experimental procedure similar to \cite{moalla2024no}, where all healthy runs use five optimization epochs per rollout, which allowed the policies to learn efficiently across all scenarios. We primarily focus on runs that remained stable (healthy) unless explicitly stated otherwise. For intentionally collapsed runs, increasing this value to 10 causes the policy to be optimized further away from the behavior policy that collected each rollout, ultimately inducing a trust-region collapse~\citep{moalla2024no}. 

As our main alternative to shallow MLPs, we implement a custom SimBa v1 \cite{lee2025simba} variant adapted for PPO, incorporating observation normalization and Pre-LN residual blocks, check Appendix~\ref{app:simba}. We selected SimBa for its architectural mechanisms intended to preserve representational capacity during training, which is clear by the wider parameter-count range with comparable returns. Nevertheless, SimBa must not be interpreted as a winner in the return sense, since both architectures can achieve comparable results under similar conditions. Both architectures were evaluated at 5 depths in the operating region, and the performance remained comparable for our analysis (Fig.~\ref{fig:mp_simba_extended}).

We conducted our real-world experiments on the Boston Dynamics Spot platform (Fig.~\ref{fig:spot}), using sim-to-real optimizations built upon the work of  ~\cite{zhu2025versatile}, with full experimental details provided in Appendix~\ref{append:real-word-experiments}. Our generalization efforts extend to evaluations on the ANYmal-D \cite{hutter2016anymal} and Unitree H1 \cite{unitree_h1} platforms, with the addition of perceptive locomotion over rough terrain. 

We rigorously run five independent seeds per hyperparameter configuration and report mean curves, unless specified otherwise. Our phase analyses focus on Spot walking on flat terrain, where we train three sizes of each family and evaluate it offline, measuring every policy in the same standardized observation scale for direct comparison (Appendix~\ref{app:rank}). Since all results are highly sensitive to learning rate ($lr$), we performed a sweep to ensure our results are comparable near their respective best observed performance; note that, in many cases, the same $lr$ was optimal for similar parameter counts. Additionally, for SimBa we implement per-configuration branch scaling in the residual path to improve training stability and mitigate $lr$ sensitivity \cite{zhang2022stabilize,bordelon2024depthwise, yang2022tensorprogramsvtuning}. 

\begin{figure}[t]
    \centering
    \includegraphics[width=\linewidth]{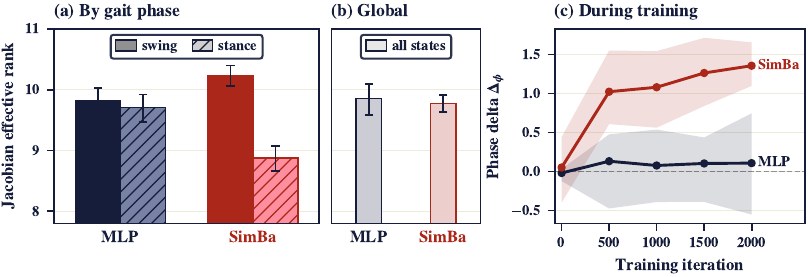}
    \caption{\textbf{Phase conditioning reveals differences hidden by global Jacobian rank.} (a) Final swing-dominant (solid) and stance-dominant (hatched) ranks. (b) Global ranks are similar (SimBa: $9.77$; MLP: $9.85$). Bars show family means with 95\% policy-bootstrap confidence intervals. (c) Swing-minus-stance phase difference $\Delta_\phi$ during training. Evaluation uses shared standardized settings (Appendix~\ref{app:rank}).}
    \label{fig:phase}
\end{figure}


\begin{figure}[t]
    \centering

    \begin{minipage}[t]{0.49\linewidth}
        \centering
        \includegraphics[width=\linewidth]{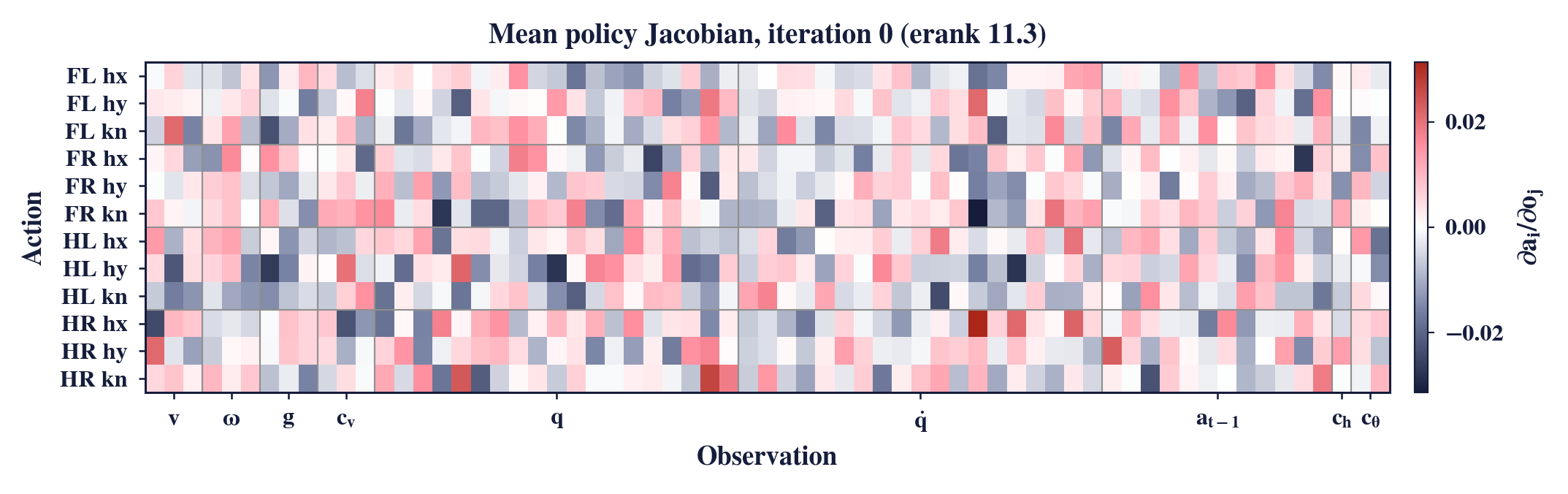}

        \small (a) Untrained
    \end{minipage}
    \hfill
    \begin{minipage}[t]{0.49\linewidth}
        \centering
        \includegraphics[width=\linewidth]{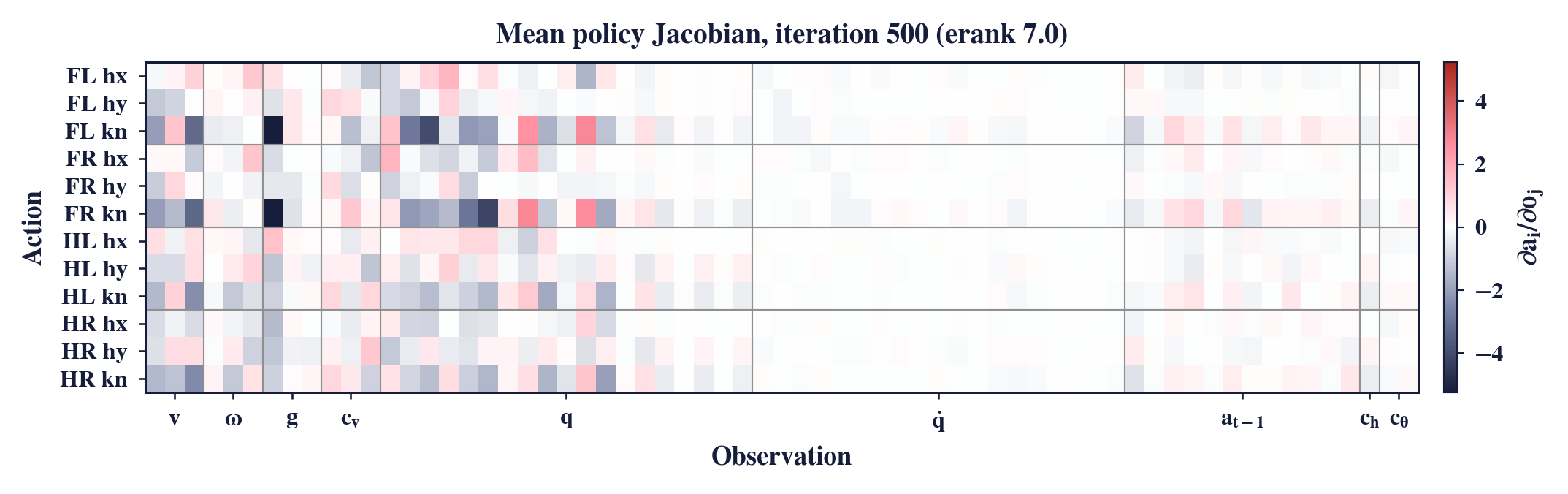}

        \small (b) Trained
    \end{minipage}

    \caption{\textbf{Jacobian matrices} for the SimBa policy in Spot environment. 
    (a) Untrained policy exhibiting near-random structure. 
    (b) Trained policy showing emergent input-output structure.}
    \label{fig:jacs}
\end{figure}

\begin{wrapfigure}{b}{0.47\columnwidth}
  \centering
  \vspace{-1em}
  \includegraphics[width=\linewidth]{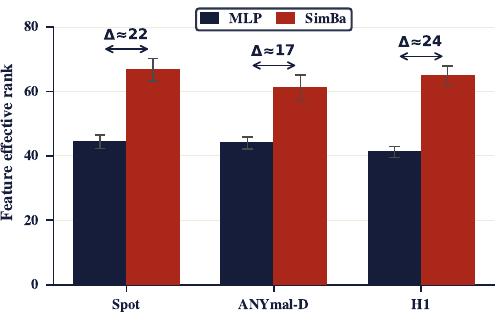}
  \caption{\textbf{Global feature effective rank} for MLP and SimBa on the flat-terrain simulation tasks for Spot, ANYmal-D, and H1. SimBa consistently keeps a $\Delta\approx20$ divergence over MLPs. Bars show mean $\pm$ spread over runs.}
  \label{fig:cross}
  \vspace{-1em}
\end{wrapfigure}

\paragraph{Conditioning rank on gait phase exposes a swing-dominant split.}

To understand how the learned policy responds during locomotion, we examine its Jacobian across gait phases. As described in Section~\ref{sec:background}, the Jacobian maps small changes in observations to changes in the action mean, while its effective rank describes how broadly this response is distributed across independent directions. With this, the jacobian lets us ask whether the policy responds differently when fewer or more feet support the body.

During training, the Jacobian develops a more structured pattern, consistent with prior observations on learned representations \citep{huh2021low}. Figure~\ref{fig:jacs} illustrates this change for a SimBa policy in the Spot environment, showing how training organizes the response of the actions to individual observation components.

As our first and main result, we found that SimBa policies consistently allocate more effective Jacobian rank to swing-dominant states than to stance-dominant states, whereas MLPs show little difference between phases (Fig.~\ref{fig:phase}a). Every SimBa policy in this comparison lies above every MLP in phase delta ($\Delta_\phi = 1.36$ vs.\ $0.11$ on average; Mann-Whitney $p < 10^{-5}$). Given that both families start from this delta close to zero (Fig.~\ref{fig:phase}c), one interpretation is that SimBa, which is designed to preserve representational capacity during training~\citep{lee2025simba}, can afford a distinct response for each phase, while the MLP responds similarly in both.

Interestingly, computing a global rank over all states, the two families become nearly indistinguishable (Fig.~\ref{fig:phase}b). This suggests that minding the gait phase should be common practice when analyzing representation health in locomotion. We also found that the magnitude of the Jacobian rank can be misleading, since which family shows the higher global rank depends on when and how the rank is measured, meanwhile the phase delta separates them consistently even during training; therefore, the main signal of this architectural difference comes from the delta between the two phases.

To see whether this architectural difference is specific to Spot, we compare the feature effective rank of both families on the flat-terrain tasks for Spot, ANYmal-D, and H1 (Fig.~\ref{fig:cross}). On all three robots, SimBa learns higher-rank representations than MLPs, by approximately 22, 17, and 24 dimensions, respectively. The two families therefore differ in their learned features across embodiments, not only in how their Jacobians change with contact on Spot, where we performed the phase analysis.

\paragraph{Higher rank policies deploy more smoothly from simulation onto hardware.}

Given the inherent relation of the Jacobian rank as a velocity quantity, we also checked how action smoothness interacts with our measurements (check Appendix~\ref{append:real-word-experiments} for details on real-world setup and metrics definitions).  In simulation on the Spot deployment environment, policies with higher Jacobian rank move more smoothly: across the architecture zoo, Jacobian rank and high-frequency joint jitter are negatively correlated (Spearman $\rho=-0.65$, $n=10$), and the higher-rank (SimBa) family shows roughly 2× lower jitter than the MLP family while matching or improving velocity tracking. On the physical Spot, SimBa policies also show lower jitter and tracking error across ten deployed policies, the same direction as in simulation. We observed the same smoothness advantage across robots, including the 19-DoF H1 biped.

We hypothesize that the implicit architectural nature, already well-known due to Simba's simplicity bias~\citep{lee2025simba}, favours low-frequency functions and, in turn, smoother actions. There is no proper artifact to link rank and smoothness directly, but we can sufficiently state that the co-occurrence phenomenon is persistent, which is why we read rank as a diagnostic of representation health that covaries with deployment quality.

\begin{figure}[t]
    \centering
    \includegraphics[width=\linewidth]{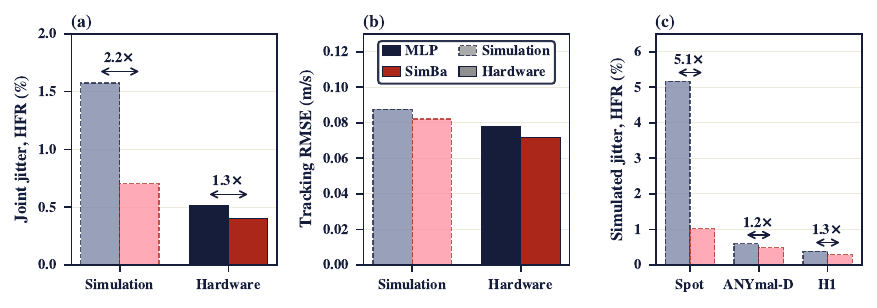}
    \caption{\textbf{Sim-to-real transfer and cross-robot generalization.} (a) Leg-joint jitter (HFR, \%) and (b) base velocity-tracking RMSE (m/s) on the Spot deployment environment, in simulation (faded) and on hardware (solid). Bars are family means over five policies each; brackets give the MLP/SimBa jitter ratio. (c) Jitter on the flat Spot, ANYmal-D, and H1 tasks (simulation only, see Appendix~\ref{app:simeval}). SimBa policies show lower jitter in every setting, with equal or better tracking.}
    \label{fig:deploy}
\end{figure}

\paragraph{Higher rank is not always better.}

\begin{figure}
    \centering
    \includegraphics[width=\linewidth]{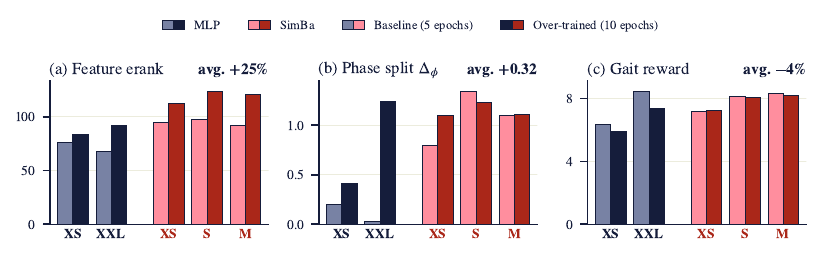}
\caption{\textbf{Rank under over-training.} Policies trained with 5 (light) or 10 (dark) PPO epochs per update, compared after 1{,}000 iterations; bars are means over five seeds and labels give the average change per configuration. (a) Inflation: feature rank grows in every configuration. (b) Reallocation: the global Jacobian rank does not change, but its phase split does, as MLPs develop a swing-dominant split while SimBa's is preserved. (c) Reward: gait reward does not improve.}
    \label{fig:collapse}
\end{figure}

Our last experiment was designed to pivot into a controlled perturbation, the closest thing to an intervention in this study. Over-training the same Spot-flat policies, by doubling the PPO epochs per update (recall: a regime known to induce trust-region collapse~\citep{moalla2024no}), changes both rank magnitude and rank structure (Figure~\ref{fig:collapse}). In this regime, during training the feature rank increases sharply (roughly 25\% for SimBa) without a comparable increase in global Jacobian rank. At the same time, the swing-dominant configuration that distinguished SimBa and MLP baselines narrowed expressively, which motivates our thesis that analyzing rank naively is not enough for fully characterizing the policy state. 

Note that the higher feature rank does not translate into a corresponding performance gain, since across these SimBa configurations, gait reward remains close to the baseline at matched iterations. Effective rank therefore shouldn't be naively maximized, but used to analyze and understand representation health. The phase-conditioned view also shows that the distinction in our first experiment depends on the training regime and not only on the architecture. In a regime associated with trust-region collapse, rank thus provides a complementary view of representation changes from only reward; an increase in rank is not, by itself, evidence of a healthier policy.

\begin{figure}
    \centering
    \includegraphics[width=0.8\linewidth]{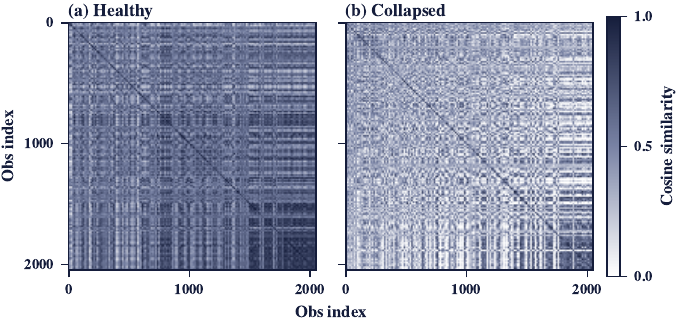}
    \caption{\textbf{Gram matrix visualization under different training regimes}. (a) Healthy training produces well-defined geometric structure, indicative of organized representation learning. (b) Collapsed training yields smooth, non-linear patterns, reflecting a loss of structure.}
    \label{fig:collapse_viz}
\end{figure}

Following \citep{huh2021low}, successful training regimes can often be interpreted as forming structured subsets in the Gram matrix. We illustrate this in Fig.~\ref{fig:collapse_viz}: under healthy training (a), clear geometric structure emerges, whereas collapsed training (b) exhibits smooth, non-linear patterns.

\section{Limitations}
\label{sec:limitations}

As with many other empirical machine learning works, the generalization of our results and the isolation of our method are admittedly not fully characterized. In the results section, we used the best sim-to-real artifacts available to validate the generalizability of our findings. Still, we consistently observed scaling trends through architectures and embodiments, and given the current quality of physical simulation, we see a clear path for future iterations. 

In methods, our experimental design and the preference for flat blind locomotion were specifically designed to create an isolated testbed in which our results would be more expressive. However, many variables still remain underexplored. In this work, we did not include a variety of architectural choices (e.g., RNNs, Transformers), normalizations, or hyper-parameter changes. We couldn't explore asymmetric decoupled actor-critic either, which is a clear trend outside flat-bind locomotion, specifically important for teacher-student pipelines and whole-body control. 

Our real-world experimentation couldn't incorporate all cross-embodiment platforms as in simulation, which may limit the generalizability of our sim-to-real findings. In general, our experimental protocol for real-world and simulation is detailed in the Appendix upon our best efforts, and we believe the current state is enough for full reproducibility and future work research.


\section{Conclusion}
\label{sec:conclusion}

We presented an analysis of PPO locomotion policies based on the effective rank, especially the policy Jacobian, and discovered that this rank becomes a useful diagnostic tool particularly when conditioned on the gait phase. In our Spot experiments, separating swing-dominant and stance-dominant states revealed differences between policies that are invisible for the global rank, while comparisons across architectures and training regimes showed why higher rank alone does not establish a healthier policy.

\paragraph{A practical recommendation.}
For researchers training locomotion policies, our recommendation is to log the effective rank of the policy Jacobian while keeping track of the robot's support condition. At regular checkpoints, collect observations together with foot-contact information and compute the rank separately for swing-dominant and stance-dominant states. Keep both measurements alongside the global rank, so that a change in their difference can be traced to what happened in each condition. Record how often the policy visits these conditions as well, including double support.

The resulting curves should be read beside reward, command-tracking error, and measurements of joint-motion smoothness. When rank changes without a corresponding improvement in behavior, this is a reason to inspect the policy more closely. It is not, by itself, a reason to favor that checkpoint. With this, representation health can be included in already moving roadmaps, which we expect to introduce new opportunities for behavior analysis. 

For readers interested in exploring these questions in their own settings, SimBa offers a useful starting point for comparison with the MLPs commonly used in locomotion, opening a practical way to ablate architectural designs.

\paragraph{Future Opportunities.}
Extending the phase-conditioned analysis to other robots, more varied terrain, and different policy architectures would help establish which observations are specific to the present setting. Controlled changes to architecture and training would also help determine which aspects of the learned response are responsible for differences in motion.

We hope that the findings from Mind the Phase provide not only a useful direction for the community, but also enough signal for further exploration of mechanistic approaches to legged locomotion.


\clearpage
\acknowledgments{
We are grateful to Francisco Affonso and Gianluca Capezzuto for insightful discussions and thoughtful suggestions that helped shape the ideas presented in this work.

This work was supported by the S\~ao Paulo Research Foundation~(FAPESP), Grant \#2025/20858-7; and by Petr\'{o}leo Brasileiro S/A - Petrobras, using resources from the ANP R\&D clause, in partnership with the University of S\~{a}o Paulo~(USP) and the Funda\c{c}\~{a}o de Apoio \`{a} F\'{\i}sica e \`{a} Qu\'{\i}mica~(FAFQ), under Cooperation Agreements \#2023/00016-6 and \#2023/00013-7.
}

\bibliography{example}  

\newpage
\appendix
\clearpage
\appendix
\renewcommand{\thetable}{A\arabic{table}}
\renewcommand{\thefigure}{A\arabic{figure}}
\setcounter{table}{0}
\setcounter{figure}{0}

\section*{Appendix}
\noindent\textbf{Contents}\par\smallskip
{\renewcommand{\arraystretch}{1.15}%
\noindent\begin{tabular}{@{}l p{0.75\textwidth} r@{}}
\hline
\ref{app:impl} & Implementation Details & \pageref{app:impl} \\
\quad\ref{app:arch} & \quad Network Architectures & \pageref{app:arch} \\
\quad\ref{app:simba} & \quad SimBa Implementation & \pageref{app:simba} \\
\quad\ref{app:baselines} & \quad PFO and LinOp Baselines & \pageref{app:baselines} \\
\quad\ref{app:ppo} & \quad PPO Hyperparameters & \pageref{app:ppo} \\
\ref{app:envs} & Simulation Environments and Training & \pageref{app:envs} \\
\quad\ref{app:isaaclab} & \quad Environment Details & \pageref{app:isaaclab} \\
\quad\ref{app:lr} & \quad Learning-Rate Selection & \pageref{app:lr} \\
\ref{app:rank} & Rank Diagnostics: Computation Details & \pageref{app:rank} \\
\quad\ref{append:contact_conditioning} & \quad Contact-State Conditioning & \pageref{append:contact_conditioning} \\
\quad\ref{app:phase_results} & \quad Phase-Conditioned Rank Results & \pageref{app:phase_results} \\
\ref{app:protocols} & Experimental Protocols & \pageref{app:protocols} \\
\quad\ref{app:sweeps} & \quad Training Cohorts & \pageref{app:sweeps} \\
\quad\ref{app:simeval} & \quad Gait Evaluation and Metrics & \pageref{app:simeval} \\
\ref{app:real} & Real-World Deployment & \pageref{app:real} \\
\quad\ref{app:hw} & \quad Hardware and Inference Pipeline & \pageref{app:hw} \\
\quad\ref{app:contact} & \quad Contact Estimation & \pageref{app:contact} \\
\quad\ref{app:hwprotocol} & \quad Protocol and Telemetry & \pageref{app:hwprotocol} \\
\ref{app:compute} & Compute Infrastructure and Reproducibility & \pageref{app:compute} \\
\ref{app:results} & Additional Results & \pageref{app:results} \\
\quad\ref{app:depth} & \quad Depth Scaling & \pageref{app:depth} \\
\quad\ref{app:collapse_results} & \quad Over-Training & \pageref{app:collapse_results} \\
\quad\ref{app:ablation} & \quad Component Ablations & \pageref{app:ablation} \\
\quad\ref{app:pfo_linop} & \quad Supplementary Baselines & \pageref{app:pfo_linop} \\
\hline
\end{tabular}}
\vspace{1.2em}

\section{Implementation Details}
\label{app:impl}

All policies were trained using the software listed in Table~\ref{tab:software} with Python~3.11. We used a fork of Isaac Lab extended with the architecture variants and with the logging and evaluation tools described in Appendices~\ref{app:rank} and~\ref{app:simeval}.

A public release version of the training code is available at:

\begin{quote}
    \url{https://github.com/EESC-LabRoM/mindthephase}
\end{quote}

Note that the infrastructure and launch trials were simplified and re-scoped to create a more stable environment for users. We plan to release reports and training logs for more in-depth insights into our results in the future.

\begin{table}[h]
\centering
\small
\renewcommand{\arraystretch}{1.3}
\setlength{\tabcolsep}{3pt}
\caption{Software stack.}
\label{tab:software}
\begin{tabular}{ll}
\hline
\textbf{Component} & \textbf{Version} \\
\hline
Simulator & Isaac Sim 5.1.0 \\
Framework & Isaac Lab 0.54.3 \\
RL library & rsl-rl-lib 5.0.1 \\
Deep learning & PyTorch 2.7.0 (CUDA 12.8) \\
Sweeps & Ray Tune 2.44.1, Optuna 4.8 \\
Tracking & MLflow, Weights \& Biases \\
Deployment & Spot SDK \\
\hline
\end{tabular}
\end{table}

\subsection{Network Architectures}
\label{app:arch}

Table~\ref{tab:arch} lists the architectures used in the main sweep.
The actor and critic use the same architecture but have separate
parameters and no shared trunk. The phase and over-training analyses
use subsets of these architectures
(Appendices~\ref{app:sweeps} and~\ref{app:overtrain}).

\begin{table}[h]
\centering
\small
\renewcommand{\arraystretch}{1.3}
\setlength{\tabcolsep}{2.4pt}
\caption{Architecture variants. Parameters (M) for the actor alone (A) and actor+critic (A+C), for a 48-D observation and a 12-D action. Counts include affine biases and layer-normalization parameters but exclude the Gaussian action-scale parameters.}
\label{tab:arch}
\begin{tabular}{@{}llllrr@{}}
\hline
 & \textbf{ID} & \textbf{Size} & \textbf{Layers / blocks} & \textbf{A} & \textbf{A+C} \\
\hline
\multirow{5}{*}{MLP}
 & MLP-XS  & XS  & $[256]\times2$          & 0.08 & 0.16 \\
 & MLP-S   & S   & $[256]\times3$          & 0.15 & 0.29 \\
 & MLP-M   & M   & $[256]\times4$          & 0.21 & 0.42 \\
 & MLP-L   & L   & $[256]\times6$          & 0.34 & 0.69 \\
 & MLP-XXL & XXL & $[512]\times4,256$      & 0.95 & 1.89 \\
\hline
\multirow{5}{*}{SimBa}
 & SimBa-XS  & XS  & $N=1$, $d=256$ & 0.54 & 1.08 \\
 & SimBa-S   & S   & $N=2$, $d=256$ & 1.07 & 2.13 \\
 & SimBa-M   & M   & $N=4$, $d=256$ & 2.12 & 4.24 \\
 & SimBa-L   & L   & $N=6$, $d=256$ & 3.17 & 6.34 \\
 & SimBa-XXL & XXL & $N=8$, $d=256$ & 4.22 & 8.45 \\
\hline
\end{tabular}
\end{table}

\subsection{SimBa Implementation}
\label{app:simba}

We follow the SimBa architecture of~\citet{lee2025simba}. A linear embedding is followed by $N$ pre-layer-normalization (Pre-LN) residual blocks, a final layer normalization (post-LN), and a linear output head,
\begin{align}
  \mathbf{h}_0
    &= W_{\text{emb}}\mathbf{x}+b_{\text{emb}}, \\
  \mathbf{h}_n
    &= \mathbf{h}_{n-1}
       + \alpha\left[
         W^{(n)}_2
         \phi\left(
           W^{(n)}_1\mathrm{LN}(\mathbf{h}_{n-1})
           + b^{(n)}_1
         \right)
         + b^{(n)}_2
       \right], \\
  \mathbf{y}
    &= W_{\text{head}}\mathrm{LN}(\mathbf{h}_N)
       + b_{\text{head}},
\end{align}
where $\phi$ is the ReLU, $W^{(n)}_1\in\mathbb{R}^{4d\times d}$ expands the hidden dimension $d$ to $4d$, and $W^{(n)}_2\in\mathbb{R}^{d\times 4d}$ projects it back to $d$.

\paragraph{Depth-stable residual scaling.} Each block's residual contribution is scaled by $\alpha = 1/\sqrt{N}$ to limit the growth of gradient magnitude with depth~\cite{zhang2022stabilize,bordelon2024depthwise}. The original SimBa formulation uses $\alpha=1$. In our implementation, the scaling is controlled by the \texttt{simba\_depth\_scale} flag, which was enabled for all SimBa runs.

\paragraph{Initialization.} The embedding and the head use orthogonal initialization with zero bias. The head gain is $0.01$ for the actor, which gives a near-zero initial action mean, and $1.0$ for the critic. Residual-block weights use Kaiming-normal initialization (fan-in, ReLU gain) with zero biases.

\paragraph{Input normalization.} SimBa policies normalize observations with running estimates of their mean and variance before the trunk (RSNorm in Table~\ref{tab:ablation}). The MLP baselines use no observation normalization. The normalizer is kept during evaluation and export. Since the two families differ in several components at once, the family comparison does not isolate the effect of normalization. Appendix~\ref{app:ablation} reports the component ablation.

\subsection{PFO and LinOp Baselines}
\label{app:baselines}

Every sweep also included two baselines that regularize the representation. Their results are reported in Appendix~\ref{app:pfo_linop}.

\paragraph{Proximal Feature Optimization (PFO).} Following Moalla et al.~\cite{moalla2024no}, PFO adds a feature-drift penalty to the PPO objective,
\begin{equation}
  \mathcal{L}_{\text{PFO}} = c_{\text{PFO}}\;\mathbb{E}_{\mathbf{x}}\big\lVert \mathbf{z}_\theta(\mathbf{x}) - \mathbf{z}_{\theta_{\text{old}}}(\mathbf{x})\big\rVert_2^2,
\end{equation}
where $\mathbf{z}$ is the actor's last pre-activation, $\theta_{\text{old}}$ is a frozen copy of the parameters taken at the start of each update, and $c_{\text{PFO}}=1$. For MLPs, $\mathbf{z}$ is the output of the last hidden linear layer, and for SimBa it is the input to the post-LN. We applied PFO to MLP-XXL and SimBa-M, giving MLP-XXL+PFO and SimBa-M+PFO.

\paragraph{Linear over-parameterization (LinOp).} LinOp replaces each square hidden layer of MLP-XXL with a product of $k$ square factors, $W = W_k\cdots W_1$, without nonlinearities between them. The function class is unchanged, but gradient descent is biased toward low-rank solutions~\cite{huh2021low}. Each factor is initialized orthogonally with gain $2^{1/(2k)}$, so all singular values of the composed matrix equal $\sqrt{2}$ and its condition number is one. We use $k\in\{2,4\}$ and call the resulting variants LinOp-2 and LinOp-4. Their learning rates and gradient clipping are given in Table~\ref{tab:ppo_params}.

\subsection{PPO Hyperparameters}
\label{app:ppo}

All tasks share the core PPO configuration from Rudin et al.~\cite{rudin2022learning}, with hyperparameters kept close to the defaults used in the original Isaac Lab tasks. Table~\ref{tab:ppo_params} lists the shared values and Table~\ref{tab:ppo_task} the task-specific ones. $K$ is the number of PPO epochs per update, set by \texttt{num\_learning\_epochs} in Isaac Lab's \texttt{RslRlPpoAlgorithmCfg}~\cite{mittal2025isaaclab}. The over-trained runs use $K=10$ instead of $5$ (Appendix~\ref{app:overtrain}). Symmetry augmentation (Sym.) is described in Appendix~\ref{app:deployenv}.

\begin{table}[h]
\centering
\small
\renewcommand{\arraystretch}{1.3}
\setlength{\tabcolsep}{2.2pt}
\caption{PPO hyperparameters shared by all runs.}
\label{tab:ppo_params}
\begin{tabular}{@{}lll@{}}
\hline
\textbf{Category} & \textbf{Parameter} & \textbf{Value} \\
\hline
\multirow{3}{*}{Rollout}
 & Steps per env. & 24 \\
 & Parallel envs. & 4{,}096 \\
 & Buffer size & 98{,}304 \\
\hline
\multirow{7}{*}{PPO}
 & Clip $\epsilon$ & 0.2 (clipped value loss) \\
 & Epochs $K$ & 5 (10 over-trained) \\
 & Mini-batches & 4 \\
 & LR schedule & Adaptive \\
 & Target KL & 0.01 \\
 & Max.\ grad.\ norm & 1.0 (0.5 LinOp) \\
 & Initial action std & 1.0 \\
\hline
\multirow{2}{*}{GAE} & Discount $\gamma$ & 0.99 \\
 & GAE $\lambda$ & 0.95 \\
\hline
\multirow{4}{*}{LR}
 & MLP (all but MLP-XXL) & $1\times10^{-3}$ \\
 & MLP (MLP-XXL) & $6\times10^{-4}$ \\
 & SimBa (all depths) & $1.5\times10^{-3}$ \\
 & LinOp ($k=2$ / $4$) & $3\times10^{-4}$ / $1.5\times10^{-4}$ \\
\hline
\end{tabular}
\end{table}

\begin{table}[h]
\centering
\small
\renewcommand{\arraystretch}{1.3}
\setlength{\tabcolsep}{2.5pt}
\caption{Task-specific PPO settings. $c_v$ is the value-loss coefficient, $c_H$ the entropy coefficient, and Sym.\ marks symmetry augmentation.}
\label{tab:ppo_task}
\begin{tabular}{lcccrc}
\hline
\textbf{Task} & $\boldsymbol{c_v}$ & $\boldsymbol{c_H}$ & \textbf{Sym.} & \textbf{Iter.} & \textbf{Seeds} \\
\hline
ANYmal-D flat      & 1.0 & 0.005  & -- & 1{,}500 & 5 \\
ANYmal-D rough     & 1.0 & 0.005  & -- & 4{,}000 & 5 \\
H1 flat            & 1.0 & 0.01   & -- & 2{,}000 & 5 \\
H1 rough           & 1.0 & 0.01   & -- & 3--4k & 5 \\
Spot flat          & 0.5 & 0.0025 & -- & 2{,}000 & 5 \\
Spot flat ($K\!=\!10$) & 0.5 & 0.0025 & -- & 1{,}000 & 5 \\
Spot deployment    & 1.0 & 0.005  & \checkmark & 10{,}000 & -- \\
\hline
\end{tabular}
\end{table}

\section{Simulation Environments and Training}
\label{app:envs}
\label{app:suite}

Table~\ref{tab:suite} lists the tasks and their identifiers throughout the appendix. The Isaac Lab velocity-tracking tasks retain their upstream definitions, except that SF uses only flat terrain patches instead of the default mixture of flat and random-rough patches. All hardware experiments use policies trained in SD, our custom Spot environment for sim-to-real transfer (Appendix~\ref{app:deployenv}).

\begin{table}[!htbp]
\centering
\scriptsize
\renewcommand{\arraystretch}{1.3}
\setlength{\tabcolsep}{3.4pt}
\caption{Simulated tasks and their identifiers (ID), with policy observation (Obs.) and action (Act.) dimensions. All tasks use joint-position targets.}
\label{tab:suite}
\begin{tabular}{@{}llllcc@{}}
\hline
\textbf{ID} & \textbf{Robot} & \textbf{Terrain} & \textbf{Task family} & \textbf{Obs.} & \textbf{Act.} \\
\hline
AF & ANYmal-D & flat  & \texttt{Isaac-Velocity-Flat-Anymal-D} & 48  & 12 \\
AR & ANYmal-D & rough & \texttt{Isaac-Velocity-Rough-Anymal-D} & 235 & 12 \\
HF & H1       & flat  & \texttt{Isaac-Velocity-Flat-H1}  & 69  & 19 \\
HR & H1       & rough & \texttt{Isaac-Velocity-Rough-H1} & 256 & 19 \\
SF & Spot     & flat  & \texttt{Isaac-Velocity-Flat-Spot} & 48 & 12 \\
SD & Spot (+arm) & flat & \texttt{Isaac-Spot-Locomotion} (deployment) & 65 & 12 \\
\hline
\end{tabular}
\end{table}

\subsection{Environment Details}
\label{app:isaaclab}

Policies receive proprioceptive observations and velocity commands, with rough-terrain policies additionally receiving a local height scan. Observation spaces, rewards, actuator models, and randomization settings follow the corresponding Isaac Lab task definitions, where their full specifications are available.

\paragraph{Spot deployment environment.}
\label{app:deployenv}

SD follows Zhu et al.~\cite{zhu2025versatile} and models Spot with its 7-DoF arm held in the stowed configuration, while the policy controls only the 12 leg joints. Its observations additionally include the arm joint state and base-height and tilt commands. Training combines locomotion rewards with mass and friction randomization, progressively introduced disturbances, and sagittal symmetry augmentation~\cite{rudin2022learning}, without manipulation rewards. The full environment configuration is publicly available in our repository.

\subsection{Learning-Rate Selection}
\label{app:lr}

\begin{figure}{h}
  \centering
  \vspace{-1em}
  \includegraphics[width=0.65\textwidth]{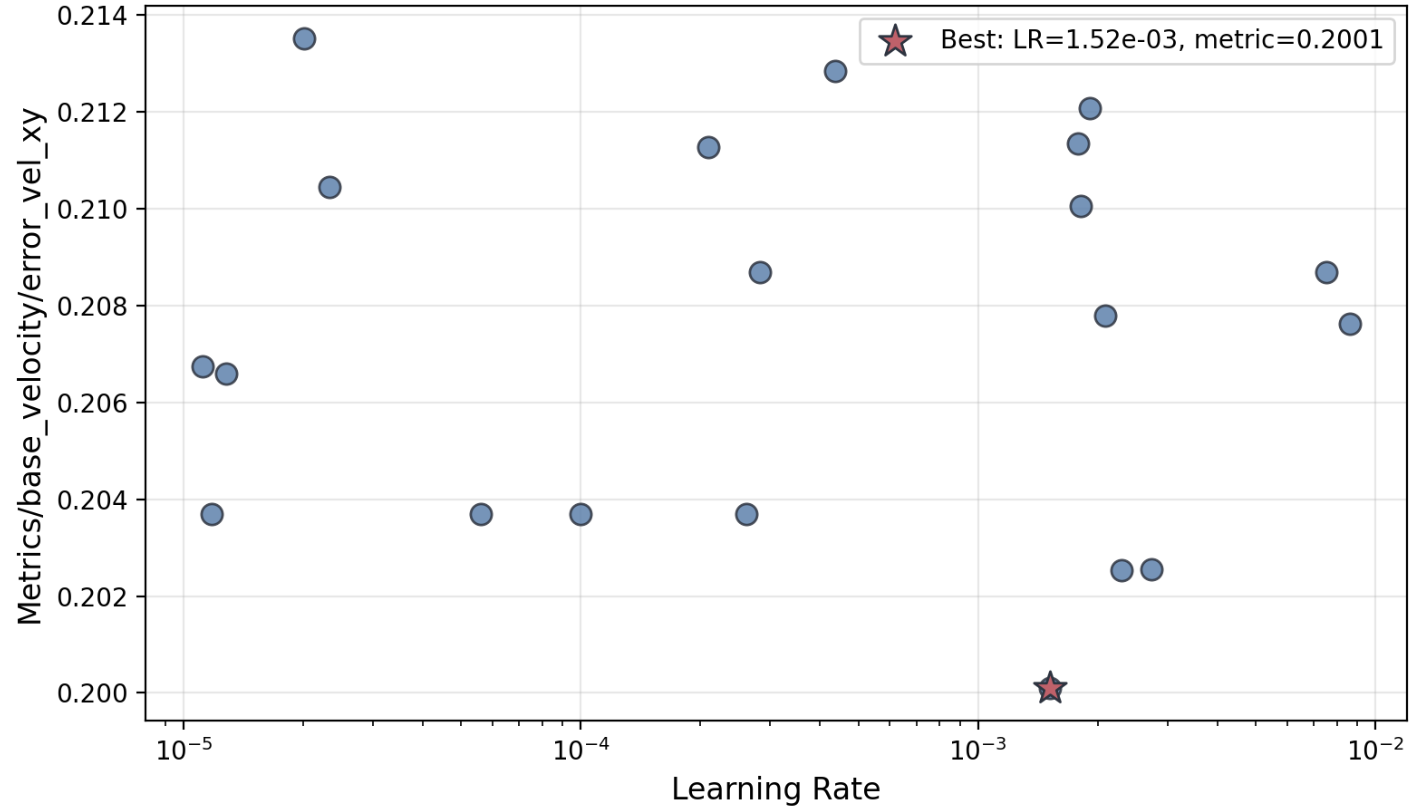}
  \caption{Bayesian optimization search for the learning rate of the SimBa architecture (SimBa-XS, ANYmal-D flat). The 20 best of the sampled points are shown.}
  \label{fig:sweep}
  \vspace{-1em}
\end{figure}

All learning rate ($lr$) values were chosen by Bayesian optimization via Optuna's Tree-Parzen Estimator (TPE) through Ray Tune, the initial $lr$ values are listed in Table~\ref{tab:ppo_params}. For instance, the smaller MLP variants use  $10^{-3}$, while MLP-XXL uses $6\times10^{-4}$. For values with less than an order of magnitude of change, we kept the optimal setting of the smaller variant. Fig.~\ref{fig:sweep} shows the 20 best trials of the
SimBa-XS sweep on ANYmal-D flat; in practice, the range
was refined iteratively.
\section{Rank Diagnostics: Computation Details}
\label{app:rank}

All diagnostics are computed on the actor network, with effective rank defined by Eq.~(3) using $\xi=10^{-10}$. Feature, Gram, and weight ranks are logged during training, together with the global Jacobian rank used in Appendix~\ref{app:simeval}. The Jacobian results in Figs.~\ref{fig:phase} and~\ref{fig:collapse} are recomputed offline using the procedure below.

\paragraph{Feature, Gram, and weight ranks.}
The feature matrix $F_\theta(X)\in\mathbb{R}^{B\times d}$ contains the inputs to the actor's final action-mean layer: the last hidden activations for MLPs and the residual-stream representations after the final layer normalization for SimBa. Every $100$ iterations, we compute its effective rank and PCA-99 rank on the current observations of all $4{,}096$ environments. PCA-99 rank is the smallest number of singular values accounting for $99\%$ of the squared Frobenius norm. Every $1{,}000$ iterations, we also compute the effective rank of the cosine-similarity Gram matrix $G=\hat F\hat F^\top$ on a fixed pool of $2{,}048$ observations, and of each actor weight matrix. LinOp layers use the collapsed product $W_k\cdots W_1$.

\paragraph{Policy Jacobian rank.}
For each state, we compute
$J_\theta(x)=\partial\boldsymbol{\mu}_\theta(x)/\partial x$
using \texttt{torch.func.vmap(jacrev)} with the actor in evaluation mode. The online logger evaluates $64$ states from the fixed pool every $1{,}000$ iterations, recording the effective rank of their mean Jacobian and the mean and standard deviation of their individual ranks. These derivatives use network-input coordinates: normalized observations for SimBa and raw observations for MLPs. Online contact-conditioned logs are used only for monitoring; all reported phase comparisons use fresh offline estimates from the same checkpoint.

\subsection{Contact-State Conditioning}
\label{append:contact_conditioning}

We condition ranks on instantaneous foot contact, defined in simulation by a vertical contact force $F_{z,f}>1\,\mathrm{N}$, without temporal averaging or hysteresis. For quadrupeds, states with zero or one contacting foot form the \textit{swing} group, while those with three or four form the \textit{stance} group. Two-contact states are excluded from the swing--stance comparison but included in the global estimate. These groups describe whole-body support, without identifying gait type, individual-leg phase, or progress through a stride. Hardware contact estimation is described in Appendix~\ref{app:contact}.

\paragraph{Offline sampling and estimation.}
We evaluate checkpoints $\{0,500,1000,1500,1999\}$ for the 5-epoch runs and $\{0,250,500,750,999\}$ for the over-trained runs; checkpoint $0$ follows the first PPO update. Each policy runs deterministically for $550$ control steps in its training environment with $4{,}096$ robots, retaining the training commands and randomization. After $300$ warm-up steps, observations and contacts are recorded every fifth step, yielding $204{,}800$ states before excluding those within $50$ steps of a reset.

For each contact group $p$, we form a random pool of at most $2{,}000$ valid states and repeatedly sample $n=64$ states without replacement to compute
\begin{equation}
    \bar J_{\theta,p}
    = \frac{1}{n}\sum_{i\in\mathcal I_p}J_\theta(x_i).
    \label{eq:phase_jac}
\end{equation}
Group ranks are the medians of $\operatorname{erank}(\bar J_{\theta,p})$ over $1{,}000$ draws. The phase difference is
\begin{equation}
    \Delta_\phi =
    \operatorname{median}_{b=1,\ldots,1000}
    \left[
    \operatorname{erank}(\bar J_{\theta,\mathrm{swing}}^{(b)})
    -
    \operatorname{erank}(\bar J_{\theta,\mathrm{stance}}^{(b)})
    \right],
    \label{eq:phase_gap}
\end{equation}
where the two groups are sampled independently and paired by draw index. Global rank uses the same estimator over all contact states.

\paragraph{Shared observation coordinates.}
Offline Jacobians are differentiated with respect to raw observations, including the observation normalizer, and expressed in shared coordinates as
$J_\theta\operatorname{diag}(\boldsymbol{\sigma}_{\mathrm{ref}})$.
The reference contains the per-dimension population standard deviations from final-checkpoint rollouts of the phase-analysis cohort, using rollout seed $0$ and the same reset filter. Weighted moments give equal weight to configurations, runs within each configuration, and valid states within each run; standard deviations are floored at $10^{-6}$. This reference is fixed across checkpoints, the closer-capacity comparison, and the over-training analysis.

\subsection{Phase-Conditioned Rank Results}
\label{app:phase_results}

The comparison includes five independently trained policies for each of MLP-XS, MLP-S, MLP-XXL, SimBa-XS, SimBa-S, and SimBa-M. At the final checkpoint, the mean phase difference is $1.36$ for SimBa and $0.11$ for MLP, with every sampled SimBa policy exceeding every sampled MLP policy (two-sided exact Mann--Whitney $p=1.3\times10^{-8}$, Cliff's $d=1.00$). Their mean global ranks are nevertheless similar, $9.77$ and $9.85$, respectively ($p=0.46$, asymptotic), showing how pooled ranks can obscure differences between contact groups.

\begin{figure}[t]
  \centering
  \includegraphics[width=\linewidth]{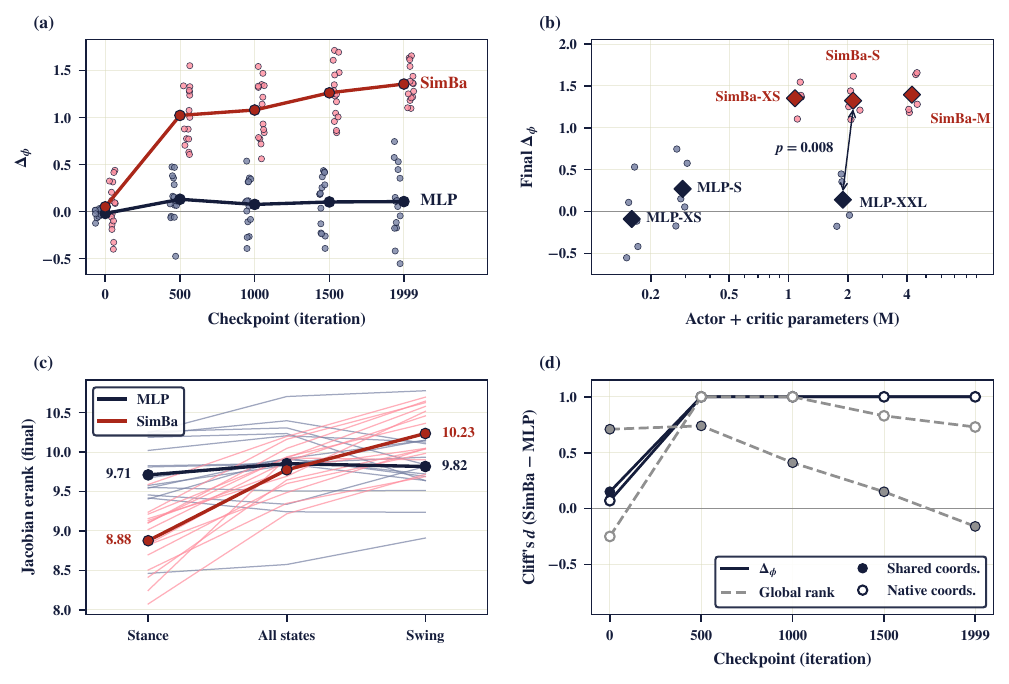}
  \caption{Phase-conditioned rank on Spot flat. Light markers and thin lines show individual policies; dark markers and thick lines show means. (a)~Phase difference through training. (b)~Final phase difference versus actor-and-critic parameter count, with configuration means as diamonds and the closer-capacity pair marked by an arrow. (c)~Final stance, global, and swing ranks. (d)~Cliff's $d$ for SimBa minus MLP in shared and native coordinates.}
  \label{fig:phase_app}
\end{figure}

The separation is also present in the closest-capacity pair, MLP-XXL and SimBa-S, with $1.89$ and $2.13$ million actor-and-critic parameters: their mean phase differences are $0.14$ and $1.32$ (exact $p=0.008$, $d=1.00$). This comparison controls parameter count approximately, without matching task performance. Across training, the family-average phase differences start near zero and separate by checkpoint $500$, with $d=1.00$ thereafter (Fig.~\ref{fig:phase_app}). One MLP policy lacks enough swing states at checkpoint $0$, leaving $14$ MLP policies at that checkpoint and $15$ elsewhere.

\paragraph{Sensitivity to measurement choices.}
Effective rank depends on observation scaling. The family separation persists in native network-input coordinates but is smaller and not statistically significant in raw observation units (Table~\ref{tab:phase_sensitivity}). It also persists with larger state subsets, a second rollout seed, and a stricter filter retaining states at least $200$ steps after reset with velocity-tracking error below the pooled median of valid states in the same rollout. This threshold is shared by both contact groups.

\begin{table}[!htbp]
\centering
\small
\caption{Sensitivity of the final mean phase difference, using $15$ policies per family. Tests are two-sided asymptotic Mann--Whitney tests; $d$ is Cliff's effect size for SimBa minus MLP.}
\label{tab:phase_sensitivity}
\begin{tabular}{lrrrr}
\hline
Setting & SimBa & MLP & $p$ & $d$ \\
\hline
Primary & $1.36$ & $0.11$ & $3.4\times10^{-6}$ & $1.00$ \\
Native coordinates & $1.40$ & $-0.08$ & $3.4\times10^{-6}$ & $1.00$ \\
Raw coordinates & $0.30$ & $-0.08$ & $0.23$ & $0.26$ \\
Age $\ge200$, low error & $1.24$ & $0.19$ & $7.5\times10^{-6}$ & $0.96$ \\
$n=256$ & $1.36$ & $0.11$ & $3.4\times10^{-6}$ & $1.00$ \\
Rollout seed 1 & $1.36$ & $0.09$ & $3.4\times10^{-6}$ & $1.00$ \\
\hline
\end{tabular}
\end{table}


\section{Experimental Protocols}
\label{app:protocols}

\subsection{Training Cohorts}
\label{app:sweeps}

For each of the five main simulation tasks, we trained the ten architectures in Table~\ref{tab:arch}, with PFO and LinOp as supplementary baselines, using a Ray Tune grid with independently sampled seeds. Simulator-startup failures were relaunched, and all completed runs are reported with their seed counts.

The phase analysis uses a separate Spot-flat cohort: MLP-XS, MLP-S, MLP-XXL, SimBa-XS, SimBa-S, and SimBa-M, each trained for $2{,}000$ iterations with five PPO epochs and seeds 101--105. Five earlier pilot runs were retained without relaunching. Each trained policy contributes one value, giving equal weight to the three configurations within each family.

\paragraph{Over-training comparison.}
\label{app:overtrain}
We retrained these configurations except MLP-S from scratch with ten PPO epochs for $1{,}000$ iterations, retaining the same five seeds and all other settings. Family comparisons include only configurations present in both regimes, without excluding policies based on contact occupancy.

Fig.~\ref{fig:collapse} compares baseline checkpoint $1000$ with over-trained checkpoint $999$, approximately matching environment samples while doubling optimization per update. Checkpoints are zero-based and saved after each update, so these represent $1{,}001$ and $1{,}000$ completed iterations. Table~\ref{tab:collapse} additionally includes baseline checkpoint $1999$, matching total PPO epochs and optimizer steps instead of environment samples. Matching seeds does not imply shared trajectories.

\paragraph{Component ablations.}
\label{app:ablation_setup}
On Spot flat, we removed observation normalization, Pre-LN, residual connections, post-LN, or SimBa initialization from SimBa-XS/S/M. Conversely, we added observation normalization, Pre-LN, residual connections, or the combination of Pre-LN and residual connections to MLP-XS/S/XXL. Each variant was trained once for $2{,}000$ iterations with seed $42$, applying changes to both actor and critic. All ablations use ReLU, including the MLP baselines, so comparisons are made within this campaign rather than against the ELU main-sweep MLPs. Table~\ref{tab:ablation} includes only completed full-length runs.

\subsection{Gait Evaluation and Metrics}
\label{app:simeval}

Simulation evaluation uses $64$ parallel robots at $50$\,Hz, with $2$\,s standing, $5$\,s walking forward, and $5$\,s walking backward at sagittal commands of $\pm0.5$\,m/s and zero lateral and yaw commands. Fig.~\ref{fig:deploy} uses the forward segments, identified from recorded commands using $v_x^{\mathrm{cmd}}>0.1$\,m/s. Each retained simulation segment contains $250$ samples without an episode reset.

For each segment, planar velocity-tracking error is
\begin{equation}
  \mathrm{RMSE}_{xy}
  =
  \left(
    \mathbb{E}_t
    \left\|
      \mathbf{v}^{\mathrm{cmd}}_{xy}(t)
      -
      \mathbf{v}^{\mathrm{base}}_{xy}(t)
    \right\|_2^2
  \right)^{1/2}.
\end{equation}
Joint-motion smoothness is measured through the high-frequency ratio (HFR). After subtracting each joint-position signal's temporal mean, we compute its real discrete Fourier transform $Q_j(f_k)$ and define
\begin{equation}
  \mathrm{HFR}_j
  =
  \frac{
    \sum_{f_k>10\,\mathrm{Hz}} |Q_j(f_k)|^2
  }{
    \sum_{f_k\ge0} |Q_j(f_k)|^2
  }.
\end{equation}
Zero-energy signals receive zero HFR. No tapering, spectral smoothing, or resampling is applied. Segment HFR is the mean over the $12$ quadruped leg joints or $10$ H1 leg joints, reported as a percentage; it measures joint-position motion rather than action jitter.

Simulation metrics are averaged equally over the $64$ robots to obtain one value per policy. Hardware uses the same metrics on forward-command segments lasting at least $2$\,s, with sampling frequency estimated separately for each log from its median control-step duration (approximately $55.6$\,Hz). Segments are identified within each log and pooled across repeated deployments using duration weights to obtain one value per policy.

\begin{figure}[t]
  \centering
  \includegraphics[width=0.9\linewidth]{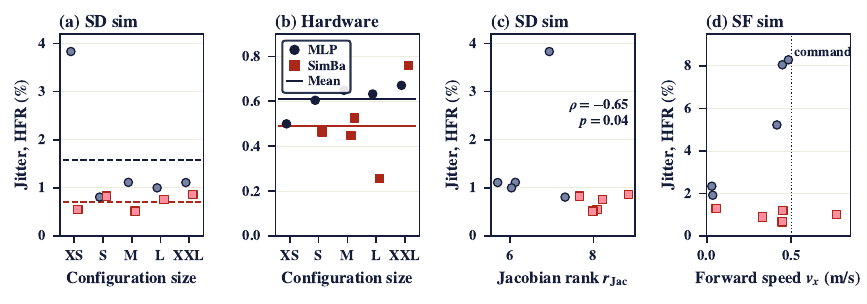}
  \caption{Per-policy forward-command results underlying Fig.~\ref{fig:deploy}. (a,b)~Simulation and hardware jitter, with family means. (c)~Global Jacobian rank versus jitter in SD simulation. (d)~Jitter versus mean forward speed, with the $0.5$\,m/s command marked by the dotted line.}
  \label{fig:deploy_policies}
\end{figure}

\paragraph{Sensitivity and scope.}
The SD simulation mean jitter ratio is sensitive to MLP-XS: it decreases from $2.23$ to $1.43$ when that policy is excluded, while the ratio of family medians is $1.46$. Hardware and H1 jitter differences are not significant at the $0.05$ level. Tracking differences are also not significant in any setting, which does not establish equivalence; H1 additionally shows that lower HFR can coincide with higher tracking error.

For the ten SD simulation policies, HFR is paired with the final global Jacobian rank logged by the corresponding training run (Appendix~\ref{app:rank}). The pooled Spearman association is negative ($\rho=-0.65$, $p=0.0425$), but weakens without MLP-XS ($\rho=-0.62$, $p=0.0769$), and within-family correlations differ in sign (MLP: $-0.10$; SimBa: $+0.40$). This association therefore does not establish that increasing rank within an architecture reduces jitter.

\section{Real-World Deployment}
\label{app:real}
\label{append:real-word-experiments}

\begin{figure}[t]
  \centering
  \includegraphics[width=\linewidth]{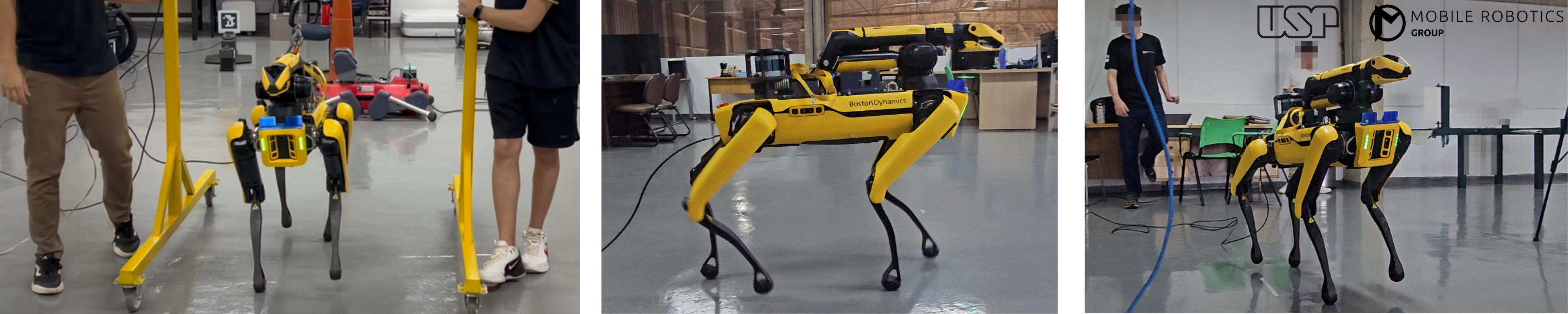}
  \caption{Hardware experiments with the deployed policies. The supplementary video shows additional runs.}
  \label{fig:hw_setup}
\end{figure}

\subsection{Hardware and Inference Pipeline}
\label{app:hw}

Inference runs in ONNX Runtime on an external desktop (Intel Core i7) connected to the robot over Ethernet. Robot states are streamed from Spot, and joint commands are sent back over gRPC through the Joint Control API of the Spot SDK. The state stream publishes joint and kinematic state at about $333$\,Hz. A new action is computed every six state packets (about $55$\,Hz, with $50$\,Hz as a lower bound), and the joint targets are streamed back with a $5$\,ms extrapolation window. End-to-end latency is $2$--$5$\,ms. The on-board joint PD loop uses $K_p=75$ and $K_d=1.875$ for the hips and $K_p=85$ and $K_d=2.0$ for the knees, and the arm joints use the default gains of the Spot SDK. Each control step logs the state arrival time, the wait time, and the ONNX inference time to verify real-time operation.

\paragraph{Observation construction.} Base linear and angular velocities come from Spot's kinematic odometry, rotated from the odometry frame into the body frame. Projected gravity is computed from the body orientation. Velocity commands are issued from a terminal interface in increments of $0.5$\,m/s. During hardware rollouts, the base roll/pitch command is held at $(0,0)$ and the base-height command at a constant value.

\subsection{Contact Estimation}
\label{app:contact}

Spot's reported contact-state estimates are logged alongside joint loads. Auxiliary hardware gait diagrams also use the signed knee-load proxy $c_f=\mathbf{1}\{\tau_{\mathrm{knee},f}>8.0\,\mathrm{N\,m}\}$. This threshold follows the sign convention of the logged loads and is not a calibrated ground-reaction-force measurement. Neither this proxy nor Spot's reported contact states are
used to compute the joint-jitter or velocity-tracking
metrics shown in Fig.~\ref{fig:deploy}. Hardware contact estimates are not used in the simulated phase-conditioned rank analysis.

\subsection{Protocol and Telemetry}
\label{app:hwprotocol}

Each policy follows the command profile of Appendix~\ref{app:simeval}, with standing and $\pm0.5$\,m/s walking. The commands were issued manually, so the number and length of segments vary between policies. The logs contain the same signals as the simulation logs that Spot provides, along with arm-joint states, commanded actions, and per-step timing. Spectral metrics use the measured rate, so the $10$\,Hz HFR cutoff is the same as in simulation. All policies were deployed with the same PD gains and a constant base-height command. Both differ from training (Appendix~\ref{app:hw}), and the deviation applies equally to both families.

The hardware group contains five MLPs and five SimBa policies. LinOp-4 and SimBa-M+PFO were also deployed and logged, but are excluded from Fig.~\ref{fig:deploy} for clearness. Repeated deployments of a checkpoint contribute to that policy's value and are not counted as additional policies. Per-policy values are shown in Fig.~\ref{fig:deploy_policies}(b).

\section{Compute Infrastructure and Reproducibility}
\label{app:compute}

\paragraph{Cluster.} Experiments ran on an in-house Ray cluster with $22$ GPUs in total ($16$ NVIDIA RTX A2000 and $6$ NVIDIA L40S). Each trial was pinned to a single GPU. Rough-terrain tasks, whose height-scan observations need more memory, ran on L40S nodes. The Spot deployment environment ran only on A2000 nodes, and flat-terrain tasks ran on either type.

\paragraph{Compute budget.}
The two main experiment-tracking projects contain $584$ training attempts with recorded runtimes, totaling about $953$ GPU-assigned hours. This total includes smoke tests and reward-development runs in addition to the main sweeps and component ablations, so it exceeds the cost of the reported runs alone. The phase-analysis and over-training cohorts contain $55$ further runs and add about $61$ GPU-assigned hours. All figures are sums of recorded run durations and not measurements of GPU utilization.

\begin{table}[!htbp]
\centering
\small
\renewcommand{\arraystretch}{1.3}
\setlength{\tabcolsep}{2.5pt}
\caption{Median wall-clock time per training run in hours, with iteration counts as in Table~\ref{tab:ppo_task}. The GPU type, workload concurrency, and training budget varied between tasks, so times are not directly comparable across rows.}
\label{tab:walltime}
\begin{tabular}{lrrrrr}
\hline
\textbf{Task} & \textbf{MLP-XS} & \textbf{MLP-XXL} & \textbf{SimBa-XS} & \textbf{SimBa-M} & \textbf{SimBa-XXL} \\
\hline
ANYmal-D flat   & 0.6 & 0.9  & 0.8 & 1.5  & 2.2 \\
ANYmal-D rough  & 3.4 & 3.5  & 3.6 & 3.9  & 4.4 \\
H1 flat         & 0.7 & 1.1  & 1.0 & 1.8  & 3.0 \\
H1 rough        & 1.0 & 1.1  & 1.1 & 1.4  & 2.3 \\
Spot flat       & 1.6 & 2.0  & 1.8 & 2.7  & 3.8 \\
Spot flat ($K\!=\!10$) & 0.6 & 0.6 & 0.6 & 0.8 & -- \\
Spot deployment & 6.8 & 10.9 & 9.6 & 18.7 & 17.9 \\
\hline
\end{tabular}
\end{table}

\begin{figure}[!tbp]
  \centering
  \includegraphics[width=\linewidth]{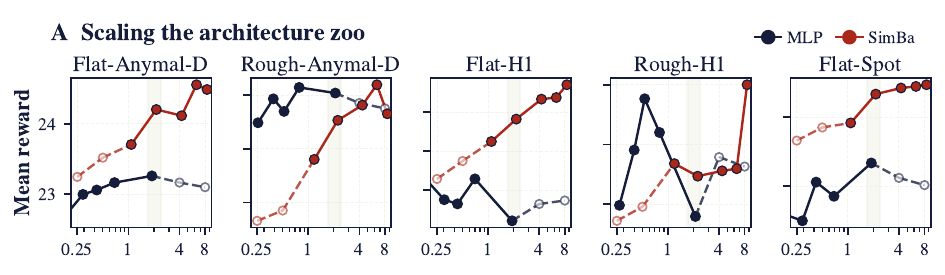}
  \caption{Performance scaling of MLP and SimBa with model capacity across tasks, over the operating range used in the paper, and a broader capacity scaled sweep. Markers show model capacity (parameters, M) on a logarithmic scale. The broader sweep shows the same qualitative scaling as the operating range.}
  \label{fig:mp_simba_extended}
\end{figure}


\section{Additional Results}
\label{app:results}

Unless stated otherwise, tables report final logged values averaged over completed training seeds ($n$). Rew.\ denotes task reward, Err.\ planar velocity error (m/s), Feat.\ actor feature effective rank, and Jac.\ global policy-Jacobian effective rank. Rewards are comparable only within each task.

\subsection{Depth Scaling}
\label{app:depth}

\begin{figure}[!tbp]
  \centering
  \includegraphics[width=\linewidth]{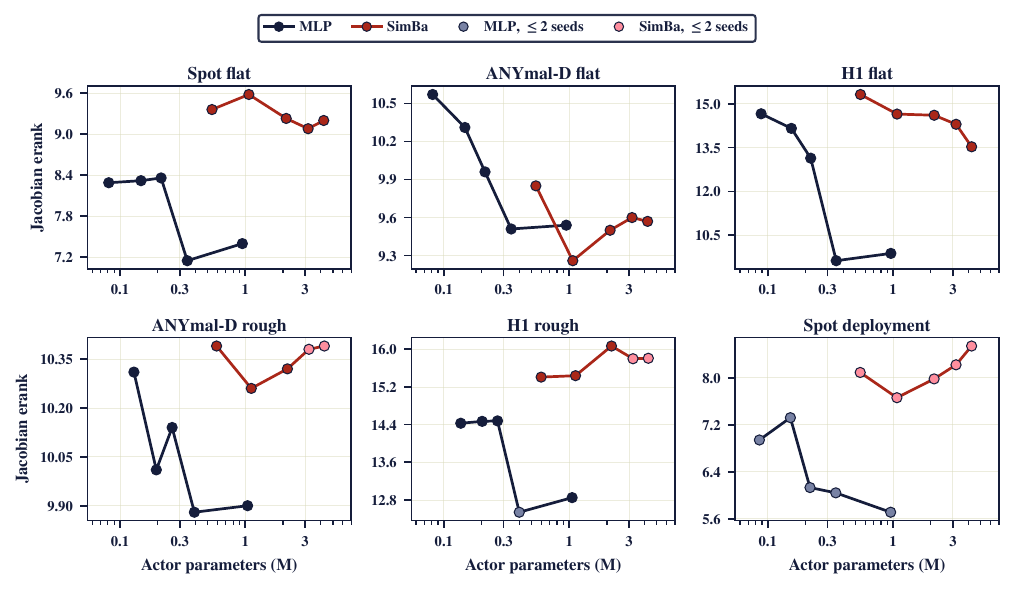}
  \caption{Global Jacobian effective rank versus actor parameter count, using each task's observation and action dimensions. }
  \label{fig:rank_depth}
\end{figure}

\begin{figure}[!tbp]
  \centering
  \includegraphics[width=\linewidth]{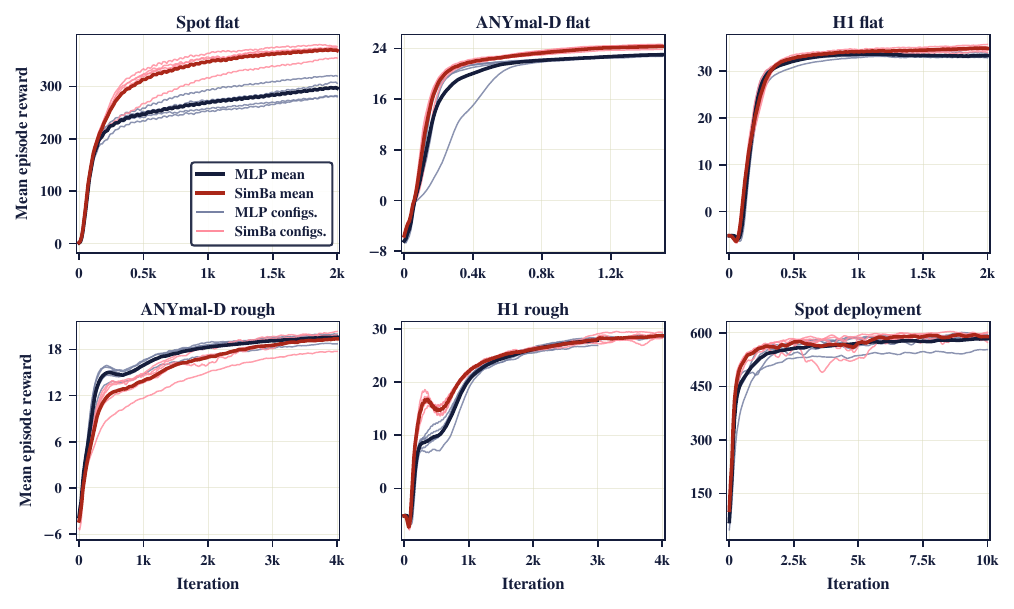}
  \caption{Training rewards: thin lines show configuration means over completed seeds; thick lines show family means over configurations. Centered moving averages use $51$ iterations ($201$ for SD). The H1-rough MLP mean ends at $3{,}000$ iterations, where three configurations stop.}
  \label{fig:training_curves}
\end{figure}

Tables~\ref{tab:res_flat} and~\ref{tab:res_rough} summarize the architecture sweep, with rank scaling and training curves shown in Figs.~\ref{fig:rank_depth} and~\ref{fig:training_curves}. SimBa consistently achieves higher mean rewards on flat terrain, while the families overlap on rough terrain. Global Jacobian rank follows different patterns across tasks, without a consistent ordering by performance. Configurations are not parameter-matched, and comparisons should account for the unequal seed counts; SD has only one run per configuration.

\begin{table}[!htbp]
\centering
\scriptsize
\renewcommand{\arraystretch}{1.05}
\setlength{\tabcolsep}{3pt}
\caption{Flat-terrain architecture sweep. Results use five seeds, except SF MLP-XS/S and SimBa-XS, AF MLP-M, and HF MLP-M, which use six. Task IDs follow Table~\ref{tab:suite}.}
\label{tab:res_flat}
\begin{tabular}{l|rrrr|rrrr|rrrr}
\hline
 & \multicolumn{4}{c|}{\textbf{Spot flat (2{,}000 iter.)}} & \multicolumn{4}{c|}{\textbf{ANYmal-D flat (1{,}500 iter.)}} & \multicolumn{4}{c}{\textbf{H1 flat (2{,}000 iter.)}} \\
\textbf{Variant} & \textbf{Rew.} & \textbf{Err.} & \textbf{Feat.} & \textbf{Jac.} & \textbf{Rew.} & \textbf{Err.} & \textbf{Feat.} & \textbf{Jac.} & \textbf{Rew.} & \textbf{Err.} & \textbf{Feat.} & \textbf{Jac.} \\
\hline
MLP-XS    & 273.2 & 1.141 & 35.0 & 8.29 & 22.62 & 0.216 & 35.7 & 10.57 & 33.35 & 0.224 & 45.0 & 14.66 \\
MLP-S     & 275.1 & 0.666 & 43.8 & 8.32 & 23.00 & 0.217 & 42.4 & 10.31 & 32.76 & 0.232 & 44.2 & 14.16 \\
MLP-M     & 302.8 & 0.457 & 47.7 & 8.36 & 23.13 & 0.208 & 48.2 & 9.96 & 32.72 & 0.244 & 42.1 & 13.14 \\
MLP-L     & 292.6 & 1.038 & 47.2 & 7.15 & 23.16 & 0.197 & 48.6 & 9.51 & 33.29 & 0.230 & 41.3 & 9.63 \\
MLP-XXL   & 316.7 & 0.418 & 48.6 & 7.40 & 23.26 & 0.201 & 45.3 & 9.54 & 32.22 & 0.252 & 33.7 & 9.88 \\
\hline
SimBa-XS  & 348.8 & 0.738 & 60.6 & 9.36 & 23.71 & 0.211 & 72.0 & 9.85 & 34.25 & 0.242 & 62.2 & 15.32 \\
SimBa-S   & 366.6 & 0.389 & 74.5 & 9.58 & 24.20 & 0.204 & 63.8 & 9.26 & 34.82 & 0.251 & 73.1 & 14.65 \\
SimBa-M   & 371.0 & 0.414 & 70.2 & 9.23 & 24.12 & 0.199 & 66.9 & 9.50 & 35.33 & 0.243 & 66.7 & 14.61 \\
SimBa-L   & 372.2 & 0.419 & 66.2 & 9.08 & 24.56 & 0.193 & 51.5 & 9.60 & 35.38 & 0.245 & 60.4 & 14.30 \\
SimBa-XXL & 373.4 & 0.467 & 61.5 & 9.20 & 24.49 & 0.194 & 51.5 & 9.57 & 35.70 & 0.235 & 62.2 & 13.53 \\
\hline
\end{tabular}
\end{table}

\begin{table}[!htbp]
\centering
\scriptsize
\renewcommand{\arraystretch}{1.05}
\setlength{\tabcolsep}{2.8pt}
\caption{Architecture sweep on rough terrain and in SD, with completed seed counts $n$. SD SimBa-XXL uses $2{,}048$ parallel environments.}
\label{tab:res_rough}
\begin{tabular}{l|crrrr|crrrr|rrrr}
\hline
 & \multicolumn{5}{c|}{\textbf{ANYmal-D rough (4{,}000 iter.)}} & \multicolumn{5}{c|}{\textbf{H1 rough (3{,}000--4{,}000 iter.)}} & \multicolumn{4}{c}{\textbf{SD (10{,}000 iter., $n=1$)}} \\
\textbf{Variant} & $\boldsymbol{n}$ & \textbf{Rew.} & \textbf{Err.} & \textbf{Feat.} & \textbf{Jac.} & $\boldsymbol{n}$ & \textbf{Rew.} & \textbf{Err.} & \textbf{Feat.} & \textbf{Jac.} & \textbf{Rew.} & \textbf{Err.} & \textbf{Feat.} & \textbf{Jac.} \\
\hline
MLP-XS    & 5 & 18.97 & 0.323 & 54.5 & 10.31 & 4 & 27.01 & 0.240 & 42.3 & 14.43 & 556.8 & 0.188 & 23.0 & 6.94 \\
MLP-S     & 5 & 19.86 & 0.302 & 56.5 & 10.01 & 5 & 28.20 & 0.232 & 43.0 & 14.47 & 595.5 & 0.173 & 21.2 & 7.32 \\
MLP-M     & 5 & 19.86 & 0.296 & 58.3 & 10.14 & 4 & 28.77 & 0.224 & 47.0 & 14.48 & 601.2 & 0.180 & 21.6 & 6.13 \\
MLP-L     & 5 & 19.90 & 0.298 & 62.2 & 9.88 & 2 & 28.20 & 0.220 & 46.3 & 12.54 & 595.7 & 0.187 & 19.6 & 6.04 \\
MLP-XXL   & 5 & 20.08 & 0.290 & 57.7 & 9.90  & 6 & 28.07 & 0.223 & 43.9 & 12.85 & 590.4 & 0.176 & 19.2 & 5.71 \\
\hline
SimBa-XS  & 6 & 17.58 & 0.341 & 48.9 & 10.39 & 5 & 27.68 & 0.227 & 25.9 & 15.41 & 584.4 & 0.164 & 12.7 & 8.09 \\
SimBa-S   & 5 & 19.06 & 0.315 & 62.5 & 10.26 & 4 & 27.69 & 0.231 & 43.0 & 15.44 & 608.1 & 0.166 & 11.9 & 7.66 \\
SimBa-M   & 5 & 19.63 & 0.306 & 50.2 & 10.32 & 3 & 27.56 & 0.230 & 30.0 & 16.07 & 599.1 & 0.167 & 11.4 & 7.98 \\
SimBa-L   & 2 & 20.27 & 0.293 & 62.3 & 10.38 & 2 & 28.50 & 0.230 & 33.5 & 15.80 & 585.0 & 0.158 & 10.8 & 8.22 \\
SimBa-XXL & 1 & 19.30 & 0.299 & 72.6 & 10.39 & 1 & 29.14 & 0.214 & 26.4 & 15.81 & 594.3 & 0.168 & 13.3 & 8.54 \\
\hline
\end{tabular}
\end{table}

\subsection{Over-Training}
\label{app:collapse_results}

\begin{figure}[!tbp]
  \centering
  \includegraphics[width=\linewidth]{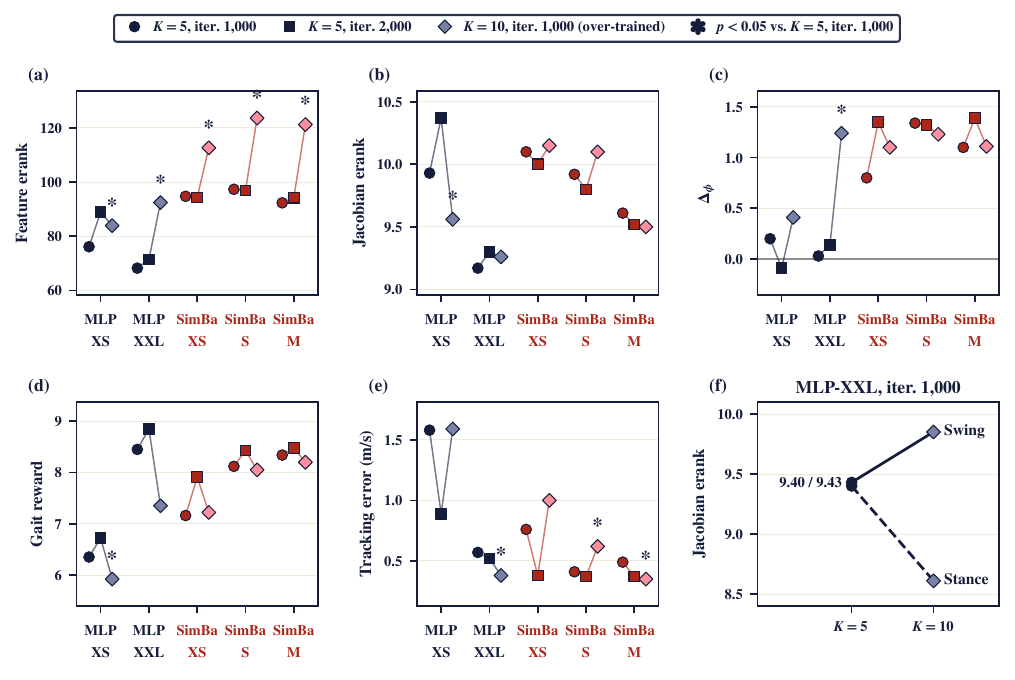}
  \caption{Over-training results from Table~\ref{tab:collapse}, with five-seed means. Asterisks indicate the same comparisons as bold table entries. (f)~MLP-XXL stance and swing Jacobian ranks at approximately $1{,}000$ iterations.}
  \label{fig:overtrain_app}
\end{figure}

Table~\ref{tab:collapse} and Fig.~\ref{fig:overtrain_app} extend the main-text comparison to the baseline endpoints, matching the total optimization budget while allowing different amounts of collected data. Over-trained policies generally have higher feature rank but lower gait reward than these endpoints, illustrating why rank alone is insufficient to assess policy health.

Here, Rew.\ is the trot-synchronization reward. Reward and tracking error are averaged over the preceding $50$ iterations, feature rank is logged at the checkpoint, and Jacobian ranks and $\Delta_\phi$ are evaluated offline (Appendix~\ref{append:contact_conditioning}). The 1,000- and 2,000-iteration labels follow the approximate training budgets; exact checkpoint indices are specified in Appendix~\ref{app:overtrain}.

For MLP-XXL, the larger phase difference combines a decrease in stance rank with an increase in swing rank. SimBa phase differences do not change significantly relative to either baseline checkpoint ($p\ge0.15$), although this does not establish equivalence. All runs remain included, including one SimBa-S policy that spends $97\%$ of its rollout in double support and has an approximately zero phase difference.

\begin{table}[!htbp]
\centering
\scriptsize
\renewcommand{\arraystretch}{1.05}
\setlength{\tabcolsep}{4pt}
\caption{Spot-flat over-training, averaged over five seeds. $K$ denotes PPO epochs per update. Bold values differ from the corresponding $K=5$, 1,000-iteration baseline (two-sided Mann--Whitney, $p<0.05$).}
\label{tab:collapse}
\begin{tabular}{@{}llrrrrrr@{}}
\hline
\textbf{Variant} & $\boldsymbol{K}$ & \textbf{Iter.} & \textbf{Feat.} & \textbf{Jac.} & $\boldsymbol{\Delta_\phi}$ & \textbf{Rew.} & \textbf{Err.} \\
\hline
\multirow{3}{*}{MLP-XS}   & 5  & 1{,}000 & 76.1 & 9.93 & 0.20 & 6.35 & 1.58 \\
                          & 5  & 2{,}000 & 88.9 & 10.37 & $-0.09$ & 6.72 & 0.89 \\
                          & 10 & 1{,}000 & \textbf{83.9} & \textbf{9.56} & 0.41 & \textbf{5.92} & 1.59 \\
\hline
\multirow{3}{*}{MLP-XXL}  & 5  & 1{,}000 & 68.2 & 9.17 & 0.03 & 8.45 & 0.57 \\
                          & 5  & 2{,}000 & 71.4 & 9.30 & 0.14 & 8.85 & 0.52 \\
                          & 10 & 1{,}000 & \textbf{92.4} & 9.26 & \textbf{1.24} & 7.35 & \textbf{0.38} \\
\hline
\multirow{3}{*}{SimBa-XS} & 5  & 1{,}000 & 94.7 & 10.10 & 0.80 & 7.16 & 0.76 \\
                          & 5  & 2{,}000 & 94.3 & 10.00 & 1.35 & 7.91 & 0.38 \\
                          & 10 & 1{,}000 & \textbf{112.6} & 10.15 & 1.10 & 7.22 & 1.00 \\
\hline
\multirow{3}{*}{SimBa-S}  & 5  & 1{,}000 & 97.3 & 9.92 & 1.34 & 8.12 & 0.41 \\
                          & 5  & 2{,}000 & 96.9 & 9.80 & 1.32 & 8.43 & 0.37 \\
                          & 10 & 1{,}000 & \textbf{123.6} & 10.10 & 1.23 & 8.05 & \textbf{0.62} \\
\hline
\multirow{3}{*}{SimBa-M}  & 5  & 1{,}000 & 92.3 & 9.61 & 1.10 & 8.34 & 0.49 \\
                          & 5  & 2{,}000 & 94.2 & 9.52 & 1.39 & 8.48 & 0.37 \\
                          & 10 & 1{,}000 & \textbf{121.2} & 9.50 & 1.11 & 8.20 & \textbf{0.35} \\
\hline
\end{tabular}
\end{table}

\subsection{Component Ablations}
\label{app:ablation}

\begin{figure}[!tbp]
    \centering
    \includegraphics[width=\linewidth]{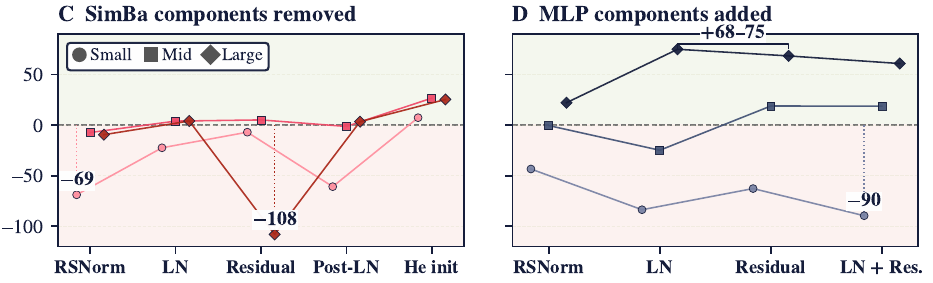}
    \caption{Spot-flat component ablations: removal from SimBa and addition to MLP.}
    \label{fig:ablation_full}
\end{figure}

Table~\ref{tab:ablation} and Fig.~\ref{fig:ablation_full} show that component effects depend on network size. Removing residual connections substantially degrades SimBa-M but has little effect on the smaller variants, while normalization changes produce mixed effects on reward and tracking. Adding components to MLPs likewise introduces trade-offs, with no single addition consistently reproducing SimBa's performance. These single-seed results are descriptive and should be compared within the ReLU ablation campaign (Appendix~\ref{app:ablation_setup}), whose MLP baselines differ from the ELU MLPs in the main sweep.

\begin{table}[!htbp]
\centering
\scriptsize
\renewcommand{\arraystretch}{1.05}
\setlength{\tabcolsep}{3pt}
\caption{Component ablations on Spot flat, with one seed and $2{,}000$ iterations per variant. All runs use ReLU; interventions affect both actor and critic.}
\label{tab:ablation}
\begin{tabular}{ll rrrr|ll rrrr}
\hline
\multicolumn{6}{c|}{\textbf{Component removal (SimBa)}} & \multicolumn{6}{c}{\textbf{Component addition (MLP)}} \\
\textbf{Base} & \textbf{Variant} & \textbf{Rew.} & \textbf{Err.} & \textbf{Feat.} & \textbf{Jac.} & \textbf{Base} & \textbf{Variant} & \textbf{Rew.} & \textbf{Err.} & \textbf{Feat.} & \textbf{Jac.} \\
\hline
\multirow{6}{*}{SimBa-XS}
 & full          & 356.7 & 0.387 & 57.8 & 9.48 & \multirow{5}{*}{MLP-XS} & base & 334.0 & 0.927 & 31.4 & 8.82 \\
 & $-$RSNorm     & 290.6 & 0.404 & 59.3 & 8.79 & & $+$RSNorm & 284.3 & 0.695 & 36.8 & 8.98 \\
 & $-$Pre-LN     & 325.0 & 0.794 & 65.8 & 9.21 & & $+$Pre-LN & 239.9 & 1.237 & 31.3 & 7.24 \\
 & $-$residual   & 364.6 & 0.411 & 73.4 & 9.56 & & $+$residual & 265.7 & 0.934 & 34.4 & 8.13 \\
 & $-$post-LN    & 301.0 & 1.906 & 92.1 & 8.49 & & $+$Pre-LN$+$res. & 235.7 & 1.648 & 32.1 & 7.31 \\
 & $-$init       & 375.2 & 0.422 & 49.9 & 10.58 & & & & & & \\
\hline
\multirow{6}{*}{SimBa-S}
 & full          & 370.0 & 0.406 & 78.9 & 9.46 & \multirow{5}{*}{MLP-S} & base & 300.3 & 0.516 & 45.0 & 8.48 \\
 & $-$RSNorm     & 358.3 & 0.401 & 73.6 & 9.12 & & $+$RSNorm & 288.0 & 1.504 & 42.7 & 8.53 \\
 & $-$Pre-LN     & 370.3 & 0.363 & 72.2 & 9.79 & & $+$Pre-LN & 253.1 & 1.131 & 40.6 & 7.04 \\
 & $-$residual   & 369.7 & 0.368 & 86.3 & 9.53 & & $+$residual & 310.0 & 1.485 & 29.5 & 9.23 \\
 & $-$post-LN    & 364.9 & 0.421 & 91.2 & 9.49 & & $+$Pre-LN$+$res. & 288.7 & 0.995 & 42.9 & 8.65 \\
 & $-$init       & 388.2 & 0.437 & 70.6 & 10.64 & & & & & & \\
\hline
\multirow{6}{*}{SimBa-M}
 & full          & 360.5 & 0.413 & 79.2 & 9.30 & \multirow{5}{*}{MLP-XXL} & base & 288.8 & 0.556 & 39.4 & 8.01 \\
 & $-$RSNorm     & 358.6 & 0.392 & 66.3 & 9.27 & & $+$RSNorm & 306.6 & 0.362 & 52.1 & 8.41 \\
 & $-$Pre-LN     & 381.3 & 0.363 & 69.8 & 9.35 & & $+$Pre-LN & 362.6 & 0.593 & 44.2 & 9.01 \\
 & $-$residual   & 255.6 & 2.065 & 6.4  & 4.91 & & $+$residual & 340.8 & 0.635 & 39.4 & 9.09 \\
 & $-$post-LN    & 373.6 & 0.357 & 95.0 & 9.58 & & $+$Pre-LN$+$res. & 344.2 & 0.451 & 45.0 & 9.16 \\
 & $-$init       & 394.9 & 0.482 & 62.4 & 10.56 & & & & & & \\
\hline
\end{tabular}
\end{table}

\subsection{Supplementary Baselines}
\label{app:pfo_linop}

Table~\ref{tab:pfo_linop} reports PFO and LinOp alongside the unregularized configurations of Tables~\ref{tab:res_flat} and~\ref{tab:res_rough}. Their effects on reward, tracking, and Jacobian rank vary by task, with limited replication on rough terrain and in SD.

\begin{table}[!htbp]
\centering
\scriptsize
\renewcommand{\arraystretch}{1.05}
\setlength{\tabcolsep}{4pt}
\caption{Supplementary PFO and LinOp results, averaged over $n$ completed runs. Task IDs follow Table~\ref{tab:suite}.}
\label{tab:pfo_linop}
\begin{tabular}{llcrrrr}
\hline
 & \textbf{Variant} & $\boldsymbol{n}$ & \textbf{Rew.} & \textbf{Err.} & \textbf{Feat.} & \textbf{Jac.} \\
\hline
\multirow{4}{*}{SF}
 & MLP-XXL+PFO  & 5 & 300.9 & 0.431 & 51.0  & 8.20 \\
 & SimBa-M+PFO  & 5 & 376.3 & 0.398 & 110.4 & 9.70 \\
 & LinOp-2      & 5 & 343.2 & 0.462 & 75.8  & 8.94 \\
 & LinOp-4      & 5 & 334.5 & 0.545 & 83.5  & 8.52 \\
\hline
\multirow{4}{*}{AF}
 & MLP-XXL+PFO  & 5 & 23.30 & 0.209 & 59.0  & 9.54 \\
 & SimBa-M+PFO  & 5 & 24.31 & 0.197 & 123.0 & 9.55 \\
 & LinOp-2      & 5 & 21.95 & 0.210 & 78.3  & 9.63 \\
 & LinOp-4      & 5 & 20.03 & 0.212 & 86.6  & 9.46 \\
\hline
\multirow{4}{*}{AR}
 & MLP-XXL+PFO  & 1 & 20.22 & 0.293 & 99.0  & 10.17 \\
 & SimBa-M+PFO  & 2 & 18.80 & 0.324 & 110.3 & 10.22 \\
 & LinOp-2      & 2 & 19.85 & 0.269 & 90.6  & 10.15 \\
 & LinOp-4      & 3 & 19.39 & 0.292 & 96.6  & 10.27 \\
\hline
\multirow{4}{*}{HF}
 & MLP-XXL+PFO  & 5 & 33.08 & 0.225 & 47.0  & 11.57 \\
 & SimBa-M+PFO  & 5 & 34.27 & 0.256 & 91.7  & 14.47 \\
 & LinOp-2      & 5 & 26.56 & 0.288 & 63.0  & 14.71 \\
 & LinOp-4      & 5 & 21.52 & 0.309 & 62.4  & 14.32 \\
\hline
\multirow{4}{*}{SD}
 & MLP-XXL+PFO  & 1 & 594.0 & 0.175 & 45.2 & 6.26 \\
 & SimBa-M+PFO  & 1 & 610.4 & 0.158 & 24.7 & 9.21 \\
 & LinOp-2      & 1 & 583.2 & 0.189 & 42.2 & 7.05 \\
 & LinOp-4      & 1 & 537.6 & 0.200 & 50.9 & 6.41 \\
\hline
\end{tabular}
\end{table}

\end{document}